\documentclass{article}

\PassOptionsToPackage{round,comma}{natbib}

     \usepackage[preprint]{neurips_2018}

\usepackage[utf8]{inputenc} 
\usepackage[T1]{fontenc}    
\usepackage{hyperref}       
\usepackage{url}            
\usepackage{booktabs}       
\usepackage{amsfonts}       
\usepackage{nicefrac}       
\usepackage{microtype}      
\usepackage{amsmath,amsthm}
\usepackage{rotating,epstopdf,graphicx,booktabs}
\usepackage{comment} 

\newtheorem{thm}{Theorem}
\newtheorem{cor}[thm]{Corollary}

\theoremstyle{definition}
\newtheorem{rem}[thm]{Remark}

\newcommand{\ier}{\textup{IER}}
\newcommand{\siid}{\sim_\textup{iid}}
\newcommand{\alg}[1]{\texttt{#1}}
\newcommand{\pv}{\textup{p-value}}

\usepackage{algorithm,algorithmic}
\newenvironment{varalgorithm}[1]
  {\floatname{algorithm}{Test}
  \algorithm\renewcommand{\thealgorithm}{#1}}
  {\endalgorithm}

\title{Practical Methods for Graph Two-Sample Testing}

\author{
  Debarghya Ghoshdastidar \\
  Department of Computer Science\\
  University of T{\"u}bingen\\
  \texttt{ghoshdas@informatik.uni-tuebingen.de} \\
  \And
  Ulrike von Luxburg \\
  Department of Computer Science\\
  University of T{\"u}bingen \\
  Max Planck Institute for Intelligent Systems \\
  \texttt{luxburg@informatik.uni-tuebingen.de} \\
}

\makeatletter\let\inserttitle\@title\makeatother

\begin{document}

\maketitle

\begin{abstract}
  Hypothesis testing for graphs has been an important tool in applied research fields for more than two decades, and still remains a challenging problem as one often needs to draw inference from few replicates of large graphs.
  Recent studies in statistics and learning theory have provided some theoretical insights about such high-dimensional graph testing problems, but the practicality of the developed theoretical methods remains an open question. 
  
  In this paper, we consider the problem of two-sample testing of large graphs. We demonstrate the practical merits and limitations of existing theoretical tests and their bootstrapped variants.
  We also propose two new tests based on asymptotic distributions. We show that these tests are computationally less expensive and, in some cases, more reliable than the existing methods.
\end{abstract}

\section{Introduction}
Hypothesis testing is one of the most commonly encountered statistical problems that naturally arises in nearly all scientific disciplines.
With the widespread use of networks in bioinformatics, social sciences and other fields since the turn of the century, it was obvious that the hypothesis testing of graphs would soon become a key statistical tool in studies based on network analysis. 
The problem of testing for differences in networks arises quite naturally in various situations.
For instance, \citet{Bassett_2008_jour_JNeuroscience} study the differences in anatomical brain networks of schizophrenic patients and healthy individuals, 
whereas \citet{Zhang_2009_jour_Bioinformatics} test for statistically significant topological changes in gene regulatory networks arising from two different treatments of breast cancer. 
As \citet{Clarke_2008_jour_NatRevCancer} and \citet{Hyduke_2013_jour_MolBioSys} point out, the statistical challenge associated with network testing is the curse of dimensionality as one needs to test large graphs based on few independent samples.
\citet{Ginestet_2014_jour_FrontComputNeurosci} show that complications can also arise due to the widespread use of multiple testing principles that rely on performing independent tests for every edge.

Although network analysis has been a primary research topic in statistics and machine learning, theoretical developments related to testing random graphs have been rather limited until recent times.
Property testing of graphs has been well studied in computer science~\citep{Goldreich_1998_jour_JACM}, but
probably the earliest instances of the theory of random graph testing are the works on community detection, which use hypothesis testing to detect if a network has planted communities or to determine the number of communities in a block model~\citep{AriasCastro_2014_jour_AnnStat,Bickel_2016_jour_JRStatistSocB,Lei_2016_jour_AnnStat}.
In the present work, we are interested in the more general and practically important problem of two-sample testing:
\emph{Given two populations of random graphs, decide whether both populations are generated from the same distribution or not.}
While there have been machine learning approaches to quantify similarities between graphs for the purpose of classification, clustering etc.~\citep{Borgwardt_2005_jour_Bioinformatics,Shervashidze_2011_jour_JMLR},
the use of graph distances for the purpose of hypothesis testing is more recent~\citep{Ginestet_2017_jour_AOAS}.
Most approaches for graph testing based on classical two-sample tests are applicable in the relatively low-dimensional setting, where the population size (number of graphs) is larger than the size of the graphs (number of vertices).
However, \citet{Hyduke_2013_jour_MolBioSys} note that this scenario does not always apply because the number of samples could be potentially much smaller --- for instance, one may need to test between two large regulatory networks (that is, population size is one).
Such scenarios can be better tackled from a perspective of high-dimensional statistics as shown in~\citet{Tang_2017_jour_JCompGraphStat,Ghoshdastidar_2017_arxiv_00833}, where the authors study two-sample testing for specific classes of random graphs with particular focus on the small population size.

In this work, we focus on the framework of the graph two-sample problem considered in~\citet{Tang_2017_jour_JCompGraphStat,Ginestet_2017_jour_AOAS,Ghoshdastidar_2017_arxiv_00833}, where all graphs are defined on a common set of vertices.
Assume that the number of vertices in each graph is $n$, and the sample size of either population is $m$.
One can consider the two-sample problem in three different regimes: 
{\bf(i)} $m$ is large; {\bf(ii)} $m>1$, but much smaller than $n$; and {\bf(iii)} $m=1$.
The first setting is the simplest one, and practical tests are known in this case~\citep{Gretton_2012_jour_JMLR,Ginestet_2017_jour_AOAS}. However, there exist many application domains where already the availability of only a small population of graphs is a challenge, and large populations are completely out of bounds. 
The latter two cases of small $m>1$ and $m=1$ have been studied in~\citet{Ghoshdastidar_2017_arxiv_00833} and~\citet{Tang_2017_jour_JCompGraphStat}, where theoretical tests based on concentration inequalities have been developed and practical bootstrapped variants of the tests have been suggested.
The contribution of the present work is three-fold:
\begin{enumerate}
\item 
For the cases of $m>1$ and $m=1$, we propose new tests that are based on asymptotic null distributions under certain model assumptions and we prove their statistical consistency (Sections~\ref{sec_m_small} and~\ref{sec_m_one} respectively).
The proposed tests are devoid of bootstrapping, and hence, computationally faster than existing bootstrapped tests for small $m$.
Detailed descriptions of the tests are provided in the Appendix~\ref{sec_algo}.
\item 
We compare the practical merits and limitations of existing tests with the proposed tests (Section~\ref{sec_expt} and Appendix~\ref{sec_addexpt}).
We show that the proposed tests are more powerful and reliable than existing methods in some situations.
\item
Our aim is also to make the existing and proposed tests more accessible for applied research.
We also provide Matlab implementations of the tests.
\end{enumerate}

The present work is focused on the assumption that all networks are defined over the same set of vertices. 
This may seem restrictive in some application areas, but it is commonly encountered in other areas such as brain network analysis or molecular interaction networks, where vertices correspond to well-defined regions of the brain or protein structures.
Few works study the case where graphs do not have vertex correspondences in context of clustering~\citep{Mukherjee_2017_conf_NIPS} and testing~\citep{Ghoshdastidar_2017_conf_COLT,Tang_2017_jour_Bernoulli}. 
But, theoretical guarantees are only known for specific choices of network functions (triangle counts or graph spectra), or under the assumption of an underlying embedding of the vertices. 

\textbf{Notation.~} We use the asymptotic notation $o_n(\cdot)$ and $\omega_n(\cdot)$, where the asymptotics are with respect to the number of vertices $n$. We say $x=o_n(y)$ and $y=\omega_n(x)$ when $\lim\limits_{n\to\infty} \frac{x}{y} = 0$. 
We denote the matrix Frobenius norm by $\Vert\cdot\Vert_F$ and the spectral norm or largest singular value by $\Vert\cdot\Vert_2$.

\section{Problem Statement}
\label{sec_problem_ier}

We consider the following framework of two-sample setting.
Let $V$ be a set of $n$ vertices. 
Let $G_1,\ldots,G_m$ and $H_1,\ldots,H_m$ be two populations of undirected unweighted graphs defined on the common vertex set $V$, where each population consists of independent and identically distributed samples.
The two-sample hypothesis testing problem is as follows:
\begin{center}
\emph{Test whether $(G_i)_{i=1,\ldots,m}$ and $(H_i)_{i=1,\ldots,m}$ are generated from the same random model or not.}
\end{center}

There exist a plethora of nonparametric tests that are provably consistent for $m\to\infty$.
In particular, kernel based tests~\citep{Gretton_2012_jour_JMLR} are known to be suitable for two-sample problems in large dimensions.
These tests, in conjunction with graph kernels~\citep{Shervashidze_2011_jour_JMLR,Kondor_2016_conf_NIPS} or distances~\citep{Mukherjee_2017_conf_NIPS}, may be used to derive consistent procedures for testing between two large populations of graphs.
Such principles are applicable even under a more general framework without vertex correspondence (see~\citealp{Gretton_2012_jour_JMLR}).
However, given graphs on a common vertex set, the most natural approach is to construct tests based on the graph adjacency matrix or the graph Laplacian~\citep{Ginestet_2017_jour_AOAS}.
To be precise, one may view each undirected graph on $n$ vertices as a $\binom{n}{2}$-dimensional vector and use classical two-sample tests based on the $\chi^2$ or $T^2$ statistics~\citep{Anderson_1984_book_Wiley}.
Unfortunately, such tests require an estimate of the $\binom{n}{2}$$\times$$\binom{n}{2}$-dimensional sample covariance matrix, which cannot be accurately obtained from a moderate sample size.
For instance, \citet{Ginestet_2017_jour_AOAS} need regularisation of the covariance estimate even for moderate sized problems $(n=40,m=100)$, and it is unknown whether such methods work for brain networks obtained from a single-lab experimental setup ($m<20$).
For $m\ll n$, it is indeed hard to prove consistency results under the general two-sample framework described above since the correlation  among the edges can be arbitrary.
Hence, we develop our theory for random graphs with independent edges.
\citet{Tang_2017_jour_JCompGraphStat} show that tests derived for such graphs are also useful in practice.

We assume that the graphs are generated from the inhomogeneous Erd\H{o}s-R{\'e}nyi (\ier) model~\citep{Bollobas_2007_jour_RSA}. 
This model has been considered in the work of~\citet{Ghoshdastidar_2017_arxiv_00833} and subsumes other models studied in the context of graph testing such as dot product graphs~\citep{Tang_2017_jour_JCompGraphStat} and stochastic block models~\citep{Lei_2016_jour_AnnStat}.
Given a symmetric matrix $P\in[0,1]^{n\times n}$ with zero diagonal, a graph $G$ is said to be an IER graph with population adjacency $P$, denoted as $G\sim\ier(P)$, if its symmetric adjacency matrix $A_G\in\{0,1\}^{n\times n}$ satisfies:
\begin{equation*}
 (A_G)_{ij}\sim \textup{Bernoulli}(P_{ij}) \text{ for all } i<j, \text{ ~~and~~ } \{(A_G)_{ij} : i<j\} \text{ are mutually independent.}
\end{equation*}
For any $n$, we state the two-sample problem as follows. 
Let $P^{(n)},Q^{(n)}\in[0,1]^{n\times n}$ be two symmetric matrices.
Given $G_1,\ldots,G_m\siid \ier\left(P^{(n)}\right)$ and  $H_1,\ldots,H_m\siid \ier\left(Q^{(n)}\right)$, 
test the hypotheses
\begin{equation}
\mathcal{H}_0: P^{(n)} = Q^{(n)} \text{~~~ against~~~ } \mathcal{H}_1: P^{(n)} \neq Q^{(n)}.
\end{equation}
Our theoretical results in subsequent sections will often be in the asymptotic case as $n\to\infty$.
For this, we assume that there are two sequences of models $\left(P^{(n)} \right)_{n\geq1}$ and $\left(Q^{(n)} \right)_{n\geq1}$, and the sequences are identical under the null hypothesis $\mathcal{H}_0$.
We derive asymptotic powers of the proposed tests assuming certain separation rates under the alternative hypothesis.

\section{Testing large population of graphs \texorpdfstring{$(m\to\infty)$}{(mtoinfty)}}
\label{sec_m_large}

Before proceeding to the case of small population size, we discuss a baseline approach that is designed for the large $m$ regime ($m\to\infty$).
The following discussion provides a $\chi^2$-type test statistic for networks, which is a simplification of~\citet{Ginestet_2017_jour_AOAS} under the IER assumption.
Given the adjacency matrices $A_{G_1},\ldots,A_{G_m}$ and  $A_{H_1},\ldots,A_{H_m}$, consider the test statistic
\begin{equation}
\label{eqn_stat_chi2}
T_{\chi^2} = \sum_{i<j} \frac{\left((\overline{A}_G)_{ij} - (\overline{A}_H)_{ij} \right)^2}{ \frac{1}{m(m-1)} \sum\limits_{k=1}^m \left((A_{G_k})_{ij} - (\overline{A}_G)_{ij}\right)^2 +  \frac{1}{m(m-1)} \sum\limits_{k=1}^m \left((A_{H_k})_{ij} - (\overline{A}_H)_{ij}\right)^2} \;,
\end{equation}
where $(\overline{A}_G)_{ij} = \frac1m \sum_{k=1}^m (A_{G_k})_{ij}$.
It is easy to see that under $\mathcal{H}_0$, $T_{\chi^2} \to \chi^2\left(\frac{n(n-1)}{2}\right)$  in distribution as $m\to\infty$ for any fixed $n$.
This suggests a $\chi^2$-type test similar to~\citet{Ginestet_2017_jour_AOAS}.
However, like any classical test, no performance guarantee can be given for small $m$ and our numerical results show that such a test is powerless for small $m$ and sparse graphs.
Hence, in the rest of the paper, we consider tests that are powerful even for small $m$.

\section{Testing small populations of large graphs \texorpdfstring{$(m>1)$}{(mgtr1)}}
\label{sec_m_small}

The case of small $m>1$ for IER graphs was first studied from a theoretical perspective in~\citet{Ghoshdastidar_2017_arxiv_00833}, and the authors also show that, under a minimax testing framework, the testing problem is quite different for $m=1$ and $m>1$.
From a practical perspective, small $m>1$ is a common situation in neural imaging with only few subjects. 
The case of $m=2$ is  also interesting for testing between two individuals based on test-retest diffusion MRI data, where two scans are collected from each subject with a separation of multiple weeks~\citep{Landman_2011_jour_Neuroimage}.

Under the assumption of IER models described in Section~\ref{sec_problem_ier} and given the adjacency matrices $A_{G_1},\ldots,A_{G_m}$ and  $A_{H_1},\ldots,A_{H_m}$, \citet{Ghoshdastidar_2017_arxiv_00833} propose test statistics based on  estimates of the distances $\left\Vert P^{(n)} - Q^{(n)} \right\Vert_2$ and $\left\Vert P^{(n)} - Q^{(n)} \right\Vert_F$ up to certain normalisation factors that account for sparsity of the graphs.
They consider the following two test statistics
\begin{align}
T_{spec} &= \frac{\left\Vert \sum\limits_{k=1}^m A_{G_k} - A_{H_k} \right\Vert_2}{\sqrt{\max\limits_{1\leq i\leq n} \sum\limits_{j=1}^n \sum\limits_{k=1}^m  (A_{G_k})_{ij} +  (A_{H_k})_{ij}}}\;,  \text{~~and}
\label{eqn_stat_spec}
\\
T_{fro} &= \frac{\sum\limits_{i<j}  \left( \sum\limits_{k\leq m/2} (A_{G_k})_{ij} -  (A_{H_k})_{ij} \right)\left( \sum\limits_{k> m/2} (A_{G_k})_{ij} -  (A_{H_k})_{ij} \right)}{ \sqrt{\sum\limits_{i<j}  \left( \sum\limits_{k\leq m/2} (A_{G_k})_{ij} +  (A_{H_k})_{ij} \right)\left( \sum\limits_{k> m/2} (A_{G_k})_{ij} +  (A_{H_k})_{ij} \right)}}\;.
\label{eqn_stat_fro}
\end{align}
Subsequently, theoretical tests are constructed based on concentration inequalities: one can show that with high probability, the test statistics are smaller than some specified threshold under the null hypothesis, but they exceed the same threshold if the separation between $P^{(n)}$ and $Q^{(n)}$ is large enough.
In practice, however, the authors note that the theoretical thresholds are too large to be exceeded for moderate $n$, and recommend estimation of the threshold through bootstrapping. 
Each bootstrap sample is generated by randomly partitioning the entire population $G_1,\ldots,G_m,H_1,\ldots,H_m$ into two parts, and $T_{spec}$ or $T_{fro}$ are computed based on this random partition.
This procedure provides an approximation of the statistic under the null model.
We refer to these tests as \alg{Boot-Spectral} and \alg{Boot-Frobenius}, and show their limitations for small $m$ via simulations.
Detailed descriptions of these tests are included in Appendix~\ref{sec_algo}.

We now propose a test based on the asymptotic behaviour of $T_{fro}$ in~\eqref{eqn_stat_fro} as $n\to\infty$.
We state the asymptotic behaviour in the following result.
\begin{thm}[{\bf Asymptotic test based on $T_{fro}$}]
\label{thm_normal}
In the two-sample framework of Section~\ref{sec_problem_ier}, assume $P^{(n)},Q^{(n)}$ have entries bounded away from 1, and satisfy $\max\left\{\left\Vert P^{(n)} \right\Vert_F ,\left\Vert Q^{(n)} \right\Vert_F\right\} = \omega_n(1)$.

Under the null hypothesis,
$\lim\limits_{n\to\infty} T_{fro}$ is dominated by a standard normal random variable, and hence, for any $\alpha\in(0,1)$,
\begin{align}
\mathbb{P}\big( T_{fro} \notin [-t_\alpha,t_\alpha] \big) \leq \alpha + o_n(1),
\end{align}
where $t_\alpha = \Phi^{-1}(1-\frac\alpha2)$ is the $\frac\alpha2$ upper quantile of the standard normal distribution.

On the other hand, if $\left\Vert P^{(n)} - Q^{(n)} \right\Vert_F^2 = \omega_n\left(\frac{1}{m}\max \left\{ \left\Vert P^{(n)} \right\Vert_F,\left\Vert Q^{(n)} \right\Vert_F \right\}\right)$, then
\begin{align}
\mathbb{P}\big( T_{fro} \in [-t_\alpha,t_\alpha] \big) = o_n(1).
\end{align}
\end{thm}

The proof, given in Appendix~\ref{sec_proof}, is based on the use of the Berry-Esseen theorem~\citep{Berry_1941_jour_AMSTrans}.
Using Theorem~\ref{thm_normal}, we propose an $\alpha$-level test based on asymptotic normal dominance of $T_{fro}$.
\begin{center}
{\bf Proposed  Test} \alg{Asymp-Normal}:
 \textit{Reject the null hypothesis if $|T_{fro}| > t_\alpha$.}
\end{center}
A detailed description of this test is given in Appendix~\ref{sec_algo}.
The assumption $\left\Vert P^{(n)} \right\Vert_F, \left\Vert Q^{(n)} \right\Vert_F = \omega_n(1)$ is not restrictive since it is  quite similar to assuming that the number of edges is super-linear in $n$, that is, the graphs are not too sparse.
We note that unlike the $\chi^2$-test of Section~\ref{sec_problem_ier}, here the asymptotics are for $n\to\infty$ instead of $m\to\infty$, and hence, the behaviour under null hypothesis may not improve for larger $m$. 
The asymptotic unit power of the \alg{Asymp-Normal} test, as shown in Theorem~\ref{thm_normal}, is proved under a separation condition, which is not surprising since we have access to only a finite number of graphs.
The result also shows that for large $m$, smaller separations can be detected by the proposed test.

\begin{rem}[{\bf Computational effort}]
\label{rem_computation}
Note that 
the computational complexity for computing the test statistics in~\eqref{eqn_stat_spec} and~\eqref{eqn_stat_fro} is \emph{linear in the total number of edges in the entire population}.
However, the bootstrap tests require computation of the test statistic multiple times (equal to number of bootstrap samples $b$; we use $b=200$ in our experiments). On the other hand, the proposed test compute the statistic once, and is much faster ($\sim$200 times).
Moreover, if the graphs are too large to be stored in memory, bootstrapping requires multiple passes over the data, while the proposed test requires only a single pass.
\end{rem}

\section{Testing difference between two large graphs \texorpdfstring{$(m=1)$}{(m=1)}}
\label{sec_m_one}

The case of $m=1$ is perhaps the most interesting from theoretical perspective: the objective is to detect whether two large graphs $G$ and $H$ are identically distributed or not.
This finds application in detecting differences in regulatory networks~\citep{Zhang_2009_jour_Bioinformatics} or comparing brain networks of individuals~\citep{Tang_2017_jour_JCompGraphStat}.
Although the concentration based test using $T_{spec}$ is applicable even for $m=1$~\citep{Ghoshdastidar_2017_arxiv_00833}, bootstrapping based on label permutation is infeasible for $m=1$ since there is no scope of permuting labels with unit population size.
\citet{Tang_2017_jour_JCompGraphStat}, however, propose a concentration based test in this case and suggest a bootstrapping based on low rank assumption of the population adjacency.
\citet{Tang_2017_jour_JCompGraphStat} study the two-sample problem for random dot product graphs, which are IER graphs with low rank population adjacency matrices (ignoring the effect of zero diagonal).
This class includes the stochastic block model, where the rank equals the number of communities. 
Let $G\sim \ier\left(P^{(n)}\right)$ and $H\sim \ier\left(Q^{(n)}\right)$, and assume that $P^{(n)}$ and $Q^{(n)}$ are of rank $r$.
One defines the adjacency spectral embedding (ASE) of graph $G$ as
$X_G = U_G \Sigma_G^{1/2}$,
where $\Sigma_G\in\mathbb{R}^{r\times r}$ is a diagonal matrix containing $r$ largest singular values of $A_G$ and $U_G\in\mathbb{R}^{n\times r}$ is the matrix of corresponding left singular vectors.
\citet{Tang_2017_jour_JCompGraphStat} propose the test statistic
\begin{equation}
\label{eqn_stat_ase}
T_{ASE} = \min \left\{ \Vert X_G - X_H W\Vert_F : W\in\mathbb{R}^{r\times r}, WW^T = I\right\},
\end{equation}
where the rank $r$ is assumed to be known. The rotation matrix $W$ aligns the ASE of the two graphs.
\citet{Tang_2017_jour_JCompGraphStat} theoretically analyse a concentration based test, where the null hypothesis is rejected if $T_{ASE}$ crosses a suitably chosen threshold.
In practice, they suggest the following bootstrapping to determine the threshold (Algorithm 1 in~\citealp{Tang_2017_jour_JCompGraphStat}).
One may approximate $P^{(n)}$ by the estimated population adjacency (EPA) $\widehat{P} = X_GX_G^T$. 
More random dot product graphs can be simulated from $\widehat{P}$, and a bootstrapped threshold can be obtained by computing $T_{ASE}$ for pairs of graphs generated from $\widehat{P}$. 
%
Instead of the $T_{ASE}$ statistic, one may also use a statistic based on EPA as
\begin{equation}
\label{eqn_stat_epa}
T_{EPA} = \left\Vert \widehat{P} - \widehat{Q} \right\Vert_F. 
\end{equation}
This statistic has been used as distance measure in the context of graph clustering~\citep{Mukherjee_2017_conf_NIPS}.
We refer to the tests based on the statistics in~\eqref{eqn_stat_ase} and~\eqref{eqn_stat_epa}, and the above bootstrapping procedure by \alg{Boot-ASE} and \alg{Boot-EPA} (see Appendix~\ref{sec_algo} for detailed descriptions). We find that the latter performs better, but both tests work under the condition that the population adjacency is of low rank, and the rank is precisely known. 
Our numerical results demonstrate the limitations of these tests when the rank is not correctly known.

Alternatively, we propose a test based on the asymptotic distribution of eigenvalues that is not restricted to graphs with low rank population adjacencies. Given $G\sim \ier\left(P^{(n)}\right)$ and $H\sim \ier\left(Q^{(n)}\right)$, consider the matrix $C\in\mathbb{R}^{n\times n}$ with zero diagonal and for $i\neq j$,
\begin{equation}
\label{eqn_diffmatrix}
C_{ij} = \frac{ (A_G)_{ij} -  (A_H)_{ij}}{\sqrt{(n-1)\left(P^{(n)}_{ij}\left(1 - P^{(n)}_{ij}\right) + Q^{(n)}_{ij}\left(1 - Q^{(n)}_{ij}\right)\right)}}.
\end{equation}
We assume that the entries of $P^{(n)}$ and $Q^{(n)}$ are not arbitrarily close to 1, and define $C_{ij}=0$ when $C_{ij}=\frac00$.
We show that the extreme eigenvalues of $C$ asymptotically follow the Tracy-Widom law, which characterises the distribution of the largest eigenvalues of matrices with independent standard normal entries~\citep{Tracy_1996_jour_CommMathPhys}.
Subsequently, we show that $\Vert C\Vert_2$ is a useful test statistic.

\begin{thm}[{\bf Asymptotic test based on $\Vert C\Vert_2$}]
\label{thm_tw}
Consider the above setting of two-sample testing, and let $C$ be as defined in~\eqref{eqn_diffmatrix}. Let $\lambda_1(C)$ and $\lambda_n(C)$ be the largest and smallest eigenvalues of $C$.

Under the null hypothesis, that is, if $P^{(n)}=Q^{(n)}$ for all $n$, then 
\begin{align*}
n^{2/3}\big(\lambda_1(C)-2\big) \to TW_1
\text{~~~and~~~}
n^{2/3}\big(-\lambda_n(C)-2\big) \to TW_1
\end{align*}
in distribution as $n\to\infty$, where $TW_1$ is the Tracy-Widom law for orthogonal ensembles. Hence, 
\begin{align}
\mathbb{P}\left( n^{2/3}(\Vert C\Vert_2-2) > \tau_\alpha \right) \leq \alpha + o_n(1),
\end{align}
for any $\alpha\in(0,1)$, where $\tau_\alpha$ is the $\frac\alpha2$ upper quantile of the $TW_1$ distribution.

On the other hand, if $P^{(n)}$ and $Q^{(n)}$ are such that $\Vert \mathbb{E}[C]\Vert_2 \geq 4+\omega_n(n^{-2/3})$, then
\begin{align}
\mathbb{P}\left( n^{2/3}(\Vert C\Vert_2-2) \leq \tau_\alpha \right) = o_n(1).
\end{align}
\end{thm}

The proof, given in Appendix~\ref{sec_proof}, relies on results on the spectrum of random matrices~\citep{Erdos_2012_jour_AdvMath,Lee_2014_jour_DukeMathJ}, and have been previously used for the special case of determining the number of communities in a block model~\citep{Bickel_2016_jour_JRStatistSocB,Lei_2016_jour_AnnStat}.
If the graphs are assumed to be block models, then asymptotic power can be proved under more precise conditions on difference in population adjacencies $P^{(n)}-Q^{(n)}$ (see Appendix~\ref{sec_thm_tw_blockmodel}).
From a practical perspective, $C$ cannot be computed since $P^{(n)}$ and $Q^{(n)}$ are unknown. 
Still, one may approximate them by relying on a weaker version of Szemer{\'e}di's regularity lemma, which implies that large graphs can be approximated by stochastic block models with possibly large number of blocks~\citep{Lovasz_2012_book_AMS}.
To this end, we propose to estimate $P^{(n)}$ from $A_G$ as follows.
We use a community detection algorithm, such as normalised spectral clustering~\citep{Ng_2002_conf_NIPS}, to find $r$ communities in $G$ ($r$ is a parameter for the test).
Subsequently $P^{(n)}$ is approximated by a block matrix $\widetilde{P}$ such that if $i,j$ lie in communities $V_1,V_2$ respectively, then $\widetilde{P}_{ij}$ is the mean of the sub-matrix of $A_G$ restricted to $V_1\times V_2$. Similarly one can also compute $\widetilde{Q}$ from $A_H$. Hence, we propose a Tracy-Widom test statistic as
\begin{align}
\label{eqn_stat_tw}
T_{TW} &= n^{2/3}\left(\left\Vert \widetilde{C} \right\Vert_2-2\right),
\\
\text{where~~~~}
\widetilde{C}_{ij} &= \frac{ (A_G)_{ij} -  (A_H)_{ij}}{\sqrt{(n-1)\left(\widetilde{P}_{ij}\left(1 - \widetilde{P}_{ij}\right) + \widetilde{Q}_{ij}\left(1 - \widetilde{Q}_{ij}\right)\right)}} \text{~~~~ for all } i\neq j
\nonumber
\end{align}
and the diagonal is zero. The proposed $\alpha$-level test based on $T_{TW}$ and Theorem~\ref{thm_tw} is the following.
\begin{center}
{\bf Proposed  Test} \alg{Asymp-TW}:
 \textit{Reject the null hypothesis if $T_{TW} > \tau_\alpha$}.
\end{center}
A detailed description of the test, as used in our implementations, is given in Appendix~\ref{sec_algo}.
We note that unlike bootstrap tests based on $T_{ASE}$ or $T_{EPA}$, the proposed test uses the number of communities (or rank) $r$ only for approximation of $P^{(n)},Q^{(n)}$, and the power of the test is not sensitive to the choice of $r$.
In addition, the computational benefit of a distribution based test over bootstrap tests, as noted in Remark~\ref{rem_computation}, is also applicable in this case.

\section{Numerical results}
\label{sec_expt}

In this section, we empirically compare the merits and limitations of the tests discussed in the paper.
We present our numerical results in three groups: (i) results for random graphs for $m>1$, (ii) results for random graphs for $m=1$, and (iii) results for testing real networks.
For $m>1$, we consider four tests.
\alg{Boot-Spectral} and \alg{Boot-Frobenius} are the bootstrap tests based on $T_{spec}$~\eqref{eqn_stat_spec} and $T_{fro}$~\eqref{eqn_stat_fro}, respectively.
\alg{Asymp-Chi2} is the $\chi^2$-type test based on $T_{\chi^2}$~\eqref{eqn_stat_chi2}, which is suited for the large $m$ setting, and finally,
the proposed test \alg{Asymp-Normal} is based on the normal dominance of $T_{fro}$ as $n\to\infty$ as shown in Theorem~\ref{thm_normal}.
For $m=1$, we consider three tests.
\alg{Boot-ASE} and \alg{Boot-EPA} are the bootstrap tests based on $T_{ASE}$~\eqref{eqn_stat_ase} and $T_{EPA}$~\eqref{eqn_stat_epa}, respectively.
\alg{Asymp-TW} is the proposed test based on $T_{TW}$~\eqref{eqn_stat_tw} and Theorem~\ref{thm_tw}.
Appendices~\ref{sec_algo} and~\ref{sec_addexpt} contain descriptions of all tests and additional numerical results.\footnote{Matlab codes available at: \url{https://github.com/gdebarghya/Network-TwoSampleTesting}}

\subsection{Comparative study on random graphs for \texorpdfstring{$m>1$}{mgtr1}}
For this study, we generate graphs from stochastic block models with 2 communities as considered in~\citet{Tang_2017_jour_JCompGraphStat}.
We define $P^{(n)}$ and $Q^{(n)}$ as follows.
The vertex set of size $n$ is partitioned into two communities, each of size $n/2$. 
In $P^{(n)}$, edges occur independently with probability $p$ within each community, and with probability $q$ between two communities.
$Q^{(n)}$ has the same block structure as $P^{(n)}$, but edges occur with probability $(p+\epsilon)$ within each community.
Under the null hypothesis $\epsilon=0$ and hence $Q^{(n)}=P^{(n)}$, whereas under the alternative hypothesis, we set $\epsilon> 0$. 

\begin{figure}[t]
\centering
\begin{tabular}{cc@{}l}
\multicolumn{2}{c}{ } & \hskip5ex Under null hypothesis  \hskip4ex Under alternative hypothesis \\
\rotatebox{90}{\hskip8ex $m=4$ \hskip15ex $m=2$} & 
\rotatebox{90}{\hskip9ex Test power (null rejection rate)} & 
\includegraphics[width=0.8\textwidth]{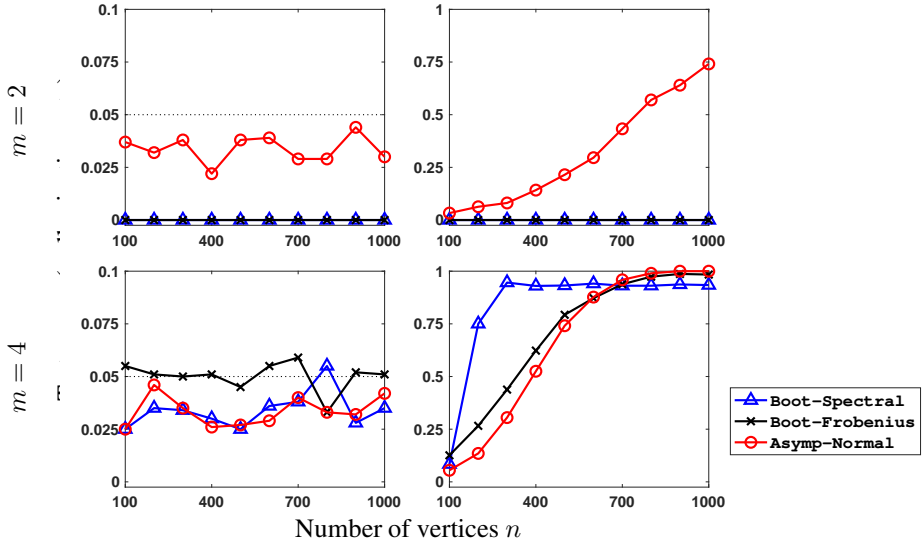}\\
& \multicolumn{2}{l}{\hskip20ex Number of vertices $n$}\\
\end{tabular}
\caption{Power of different tests for increasing number of vertices $n$, and for $m=2,4$. The dotted line for case of null hypothesis corresponds to the significance level of 5\%.}
\label{fig_1a}
\end{figure}

In our first experiment, we study the performance of different tests for varying $m$ and $n$.
We let $n$ grow from 100 to 1000 in steps of 100, and set $p = 0.1$ and $q = 0.05$. We set $\epsilon=0$ and 0.04 for null and alternative hypotheses, respectively. 
We use two values of population size, $m\in\{2, 4\}$, and fix the significance level at $\alpha = 5\%$.
Figure~\ref{fig_1a} shows the rate of rejecting the null hypothesis (test power) computed from 1000 independent runs of the experiment.
Under the null model, the test power should be smaller than $\alpha = 5\%$, whereas under the alternative model, a high test power (close to 1) is desirable.
We see that for $m=2$, only \alg{Asymp-Normal} has power while the bootstrap tests have zero rejection rate. This is not surprising as bootstrapping is impossible for $m=2$. For $m=4$, \alg{Boot-Frobenius} has a behaviour similar to \alg{Asymp-Normal} although the latter is computationally much faster.
\alg{Boot-Spectral} achieves a higher power for small $n$ but cannot achieve unit power.
\alg{Asymp-Chi2} has an erratic behaviour for small $m$, and hence, we study it for larger sample size in Figure~\ref{fig_1c} (in Appendix~\ref{sec_addexpt}).
As is expected, \alg{Asymp-Chi2} has desired performance only for $m\gg n$.

We also study the effect of edge sparsity on the performance of the tests. For this, we consider the above setting, but scale the edge probabilities by a factor of $\rho$, where $\rho = 1$ is exactly same as the above setting while larger $\rho$ corresponds to denser graphs. Figure~\ref{fig_1b} in the appendix shows the results in this case, where we fix $n=500$ and vary $\rho\in\{\frac14,\frac12,1,2,4\}$ and $m\in\{2,4,6,8,10\}$.
We again find that \alg{Asymp-Normal} and \alg{Boot-Frobenius} have similar trends for $m\geq4$. All tests perform better for dense graphs, but \alg{Boot-Spectral} may be preferred for sparse graphs when $m\geq6$.

\subsection{Comparative study on random graphs for \texorpdfstring{$m=1$}{m=1}}

\begin{figure}[t]
\centering
\begin{tabular}{cc@{}l}
\multicolumn{2}{c}{ } & \hskip4ex Under null hypothesis  \hskip5ex Under alternative hypothesis \\
\rotatebox{90}{\hskip8ex $r=4$ \hskip17ex $r=2$} & 
\rotatebox{90}{\hskip9ex Test power (null rejection rate)} & 
\includegraphics[width=0.75\textwidth]{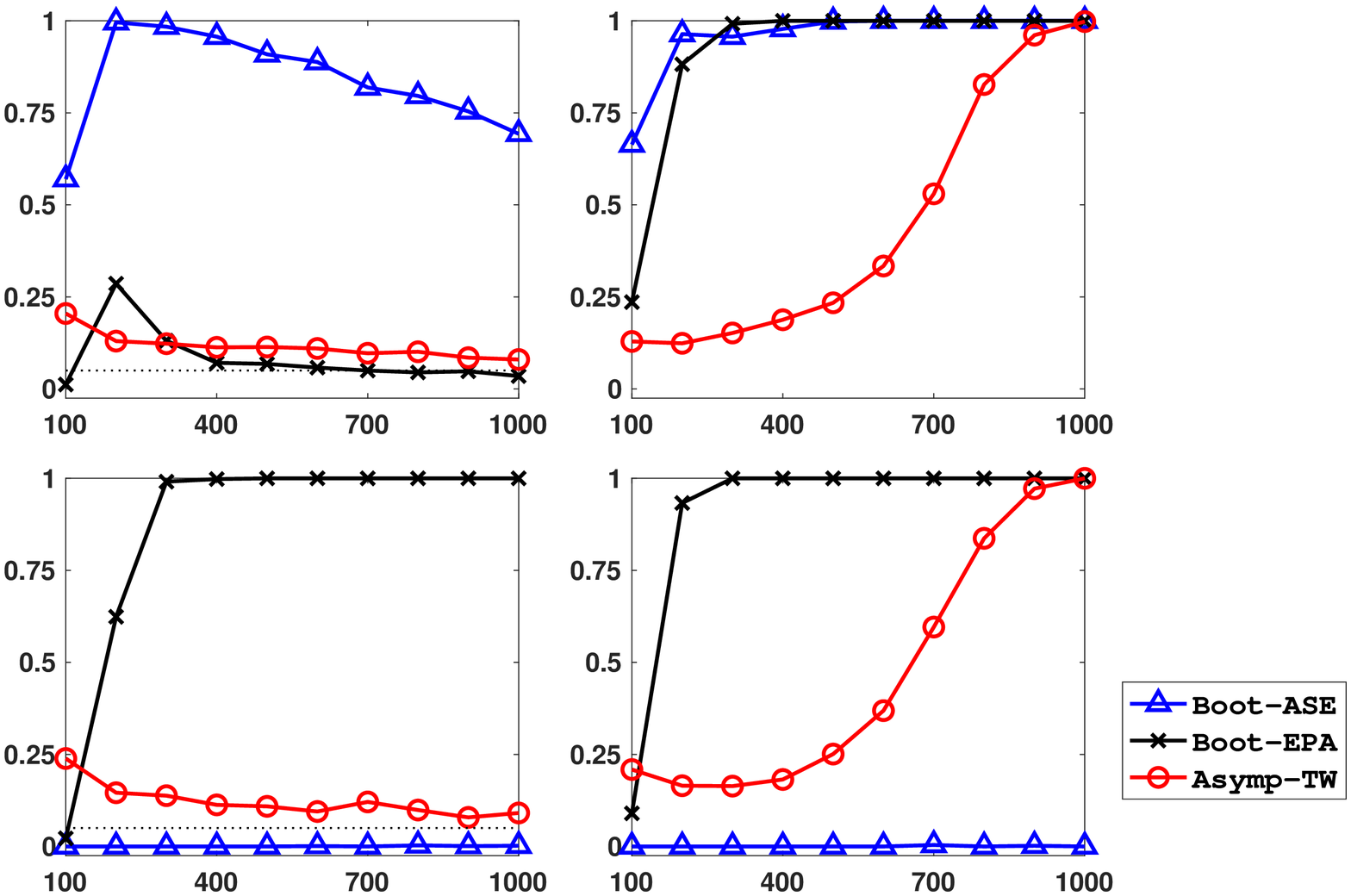}\\
& \multicolumn{2}{l}{\hskip20ex Number of vertices $n$}\\
\end{tabular}
\caption{Power of different tests with increase number of vertices $n$, and for rank parameter $r=2,4$. The dotted line under null hypothesis corresponds to the significance level of 5\%.}
\label{fig_2a}
\end{figure}

We conduct similar experiments for the case of $m=1$. 
Recall that bootstrap tests for $m=1$ work under the assumption that the population adjacencies are of low rank. This holds in above considered setting of block models, where the rank is 2. We first demonstrate the effect of knowledge of true rank on the test power. We use $r\in\{2,4\}$ to specify the rank parameter for bootstrap tests, and also as the number of blocks used for community detection step of \alg{Asymp-TW}.
Figure~\ref{fig_2a} shows the power of the tests for the above setting with $\rho=1$ and growing $n$.
We find that when $r=2$, that is, true rank is known, both bootstrap tests perform well under alternative hypothesis, and outperform \alg{Asymp-TW}, although \alg{Boot-ASE} has a high type-I error rate.
However, when an over-estimate of rank is used $(r=4)$, both bootstrap tests break down --- \alg{Boot-EPA} always rejects while \alg{Boot-ASE} always accepts --- but the performance of \alg{Asymp-TW} is robust to this parameter change.

We also study the effect of sparsity by varying $\rho$ (see Figure~5 in Appendix~\ref{sec_addexpt}).
We only consider the case $r=2$.
We find that all tests perform better in dense regime, and the rejection rate of \alg{Asymp-TW} under null is below 5\% even for small graphs.
However, the performance of both \alg{Boot-ASE} and \alg{Asymp-TW} are poor if the graphs are too sparse.
Hence, \alg{Boot-EPA} may be preferable for sparse graphs, but only if the rank is correctly known.

\subsection{Qualitative results for testing  real networks}

We use the proposed asymptotic tests to analyse two real datasets.
These experiments demonstrate that the proposed tests are applicable beyond the setting of IER graphs.
In the first setup, we consider moderate sized graphs $(n=178)$ constructed by thresholding autocorrelation matrices of EEG recordings~\citep{Andrzejak_2001_jour_PhysRevE,Dua_2017_data_uci}. The network construction is described Appendix~\ref{sec_seizure}.
Each group of networks corresponds to either epileptic seizure activity or four other resting states.
In Tables~\ref{tab_seiz_pow2}--\ref{tab_seiz_pval1} in Appendix~\ref{sec_addexpt}, we report the test powers and p-values for \alg{Asymp-Normal} and \alg{Asymp-TW}.
We find that, except for one pair of resting states, networks for different groups can be distinguished by both tests.
Further observations and discussions are also provided in the appendix.

We also study networks corresponding to peering information of autonomous systems, that is, graphs defined on the routers comprising the Internet with the edges representing \emph{who-talks-to-whom}~\citep{Leskovec_2005_conf_KDD,Leskovec_2014_data_snap}. 
The information for $n=11806$ systems was collected once a week for nine consecutive weeks, and two networks are available for each date based on two sets of information $(m=2)$.
We run  \alg{Asymp-Normal} test for every pair of dates and report the p-values in Table~\ref{tab_oregon} (Appendix~\ref{sec_oregon}).
It is interesting to observe that as the interval between two dates increase, the p-values decrease at an exponential rate, that is, the networks differ drastically according to our tests.
We also conduct semi-synthetic experiments by randomly perturbing the networks, and study the performance of \alg{Asymp-Normal} and \alg{Asymp-TW} as the perturbations increase (see Figures~6--7).
Since the networks are large and sparse, we perform the community detection step of \alg{Asymp-TW} using BigClam~\citep{Yang_2013_conf_WSDM} instead of spectral clustering. 
We infer that the limitation of \alg{Asymp-TW} in sparse regime (observed in Figure~5) could possibly be caused by poor performance of standard spectral clustering in sparse regime. 

\section{Concluding remarks}
In this work, we consider the two-sample testing problem for undirected unweighted graphs defined on a common vertex set. 
This problem finds application in various domains, and is often challenging due to unavailability of large number of samples (small $m$).
We study the practicality of existing theoretical tests, and propose two new tests based on asymptotics for large graphs (Thereoms~\ref{thm_normal} and~\ref{thm_tw}).
We perform numerical comparison of various tests, and also provide their Matlab implementations.
In the $m>1$ case, we find that \alg{Boot-Spectral} is effective for $m\geq6$, but \alg{Asymp-Normal} is recommended for smaller $m$  since it is more reliable and requires less computation.
For $m=1$, we recommend \alg{Asymp-TW} due to robustness to the rank parameter and computational advantage.
For large sparse graphs, \alg{Asymp-TW} should be used with a robust community detection step (BigClam).

One can certainly extend some of these tests to more general frameworks of graph testing.
For instance, \emph{directed graphs} can be tackled by modifying $T_{fro}$ such that the summation is over all $i,j$ and Theorem~\ref{thm_normal} would hold even in this case.
For \emph{weighted graphs}, Theorem~\ref{thm_tw} can be used if one modifies $C$~\eqref{eqn_diffmatrix} by normalising with variance of $(A_G)_{ij}-(A_H)_{ij}$.
Subsequently, these variances can be approximated again through block modelling.
For $m>1$, we believe that \emph{unequal population sizes} can be handled by rescaling the matrices appropriately, but we have not verified this.

\subsection*{Acknowledgements}
This work is supported by the German Research Foundation (Research Unit 1735) and the Institutional Strategy of the University of T{\"u}bingen (DFG, ZUK 63). 

\bibliographystyle{abbrvnat}
\bibliography{refs}

\begin{thebibliography}{34}
\providecommand{\natexlab}[1]{#1}
\providecommand{\url}[1]{\texttt{#1}}
\expandafter\ifx\csname urlstyle\endcsname\relax
  \providecommand{\doi}[1]{doi: #1}\else
  \providecommand{\doi}{doi: \begingroup \urlstyle{rm}\Url}\fi

\bibitem[Anderson(1984)]{Anderson_1984_book_Wiley}
T.~W. Anderson.
\newblock \emph{An introduction to multivariate statistical analysis}.
\newblock John Wiley and Sons, 1984.

\bibitem[Andrzejak et~al.(2001)Andrzejak, Lehnertz, Rieke, Mormann, David, and
  Elger]{Andrzejak_2001_jour_PhysRevE}
R.~G. Andrzejak, K.~Lehnertz, C.~Rieke, F.~Mormann, P.~David, and C.~E. Elger.
\newblock Indications of nonlinear deterministic and finite dimensional
  structures in time series of brain electrical activity: Dependence on
  recording region and brain state.
\newblock \emph{Physical Review E}, 64:\penalty0 061907, 2001.

\bibitem[Arias-Castro and Verzelen(2014)]{AriasCastro_2014_jour_AnnStat}
E.~Arias-Castro and N.~Verzelen.
\newblock Community detection in dense random networks.
\newblock \emph{Annals of Statistics}, 42\penalty0 (3):\penalty0 940--969,
  2014.

\bibitem[Bassett et~al.(2008)Bassett, Bullmore, Verchinski, Mattay, Weinberger,
  and Meyer-Lindenberg]{Bassett_2008_jour_JNeuroscience}
D.~S. Bassett, E.~Bullmore, B.~A. Verchinski, V.~S. Mattay, D.~R. Weinberger,
  and A.~Meyer-Lindenberg.
\newblock Hierarchical organization of human cortical networks in health and
  schizophrenia.
\newblock \emph{The Journal of Neuroscience}, 28\penalty0 (37):\penalty0
  9239--9248, 2008.

\bibitem[Berry(1941)]{Berry_1941_jour_AMSTrans}
A.~C. Berry.
\newblock The accuracy of the {Gaussian} approximation to the sum of
  independent variates.
\newblock \emph{Transactions of the American Mathematical Society}, 49\penalty0
  (1):\penalty0 122--136, 1941.

\bibitem[Bickel and Sarkar(2016)]{Bickel_2016_jour_JRStatistSocB}
P.~J. Bickel and P.~Sarkar.
\newblock Hypothesis testing for automated community detection in networks.
\newblock \emph{Journal of the Royal Statistical Society Series B: Statistical
  Methodology}, 78\penalty0 (1):\penalty0 253--273, 2016.

\bibitem[Bollobas et~al.(2007)Bollobas, Janson, and
  Riordan]{Bollobas_2007_jour_RSA}
B.~Bollobas, S.~Janson, and O.~Riordan.
\newblock The phase transition in inhomogeneous random graphs.
\newblock \emph{Random Structures and Algorithms}, 31\penalty0 (1):\penalty0
  3--122, 2007.

\bibitem[Borgwardt et~al.(2005)Borgwardt, Ong, Sch{\"o}nauer, Vishwanathan,
  Smola, and Kriegel]{Borgwardt_2005_jour_Bioinformatics}
K.~M. Borgwardt, C.~S. Ong, S.~Sch{\"o}nauer, S.~V. Vishwanathan, A.~J. Smola,
  and H.~P. Kriegel.
\newblock Protein function prediction via graph kernels.
\newblock \emph{Bioinformatics}, 21\penalty0 (Suppl 1):\penalty0 i47--56, 2005.

\bibitem[Bornemann(2010)]{Bornemann_2010_jour_MarkovProc}
F.~Bornemann.
\newblock On the numerical evaluation of distributions in random matrix theory.
\newblock \emph{Markov Processes and Related Fields}, 16:\penalty0 803--866,
  2010.

\bibitem[Clarke et~al.(2008)Clarke, Ressom, Wang, Xuan, Liu, Gehan, and
  Wang]{Clarke_2008_jour_NatRevCancer}
R.~Clarke, H.~W. Ressom, A.~Wang, J.~Xuan, M.~C. Liu, E.~A. Gehan, and Y.~Wang.
\newblock The properties of high-dimensional data spaces: Implications for
  exploring gene and protein expression data.
\newblock \emph{Nature Reviews Cancer}, 8:\penalty0 37--49, 2008.

\bibitem[Dua and Taniskidou(2017)]{Dua_2017_data_uci}
D.~Dua and K.~Taniskidou.
\newblock {UCI} machine learning repository.
\newblock \url{http://archive.ics.uci.edu/ml}, 2017.

\bibitem[Erd{\H{o}}s et~al.(2012)Erd{\H{o}}s, Yau, and
  Yin]{Erdos_2012_jour_AdvMath}
L.~Erd{\H{o}}s, H.-T. Yau, and J.~Yin.
\newblock Rigidity of eigenvalues of generalized {Wigner} matrices.
\newblock \emph{Advances in Mathematics}, 229\penalty0 (3):\penalty0
  1435--1515, 2012.

\bibitem[Ghoshdastidar et~al.(2017{\natexlab{a}})Ghoshdastidar, Gutzeit,
  Carpentier, and von Luxburg]{Ghoshdastidar_2017_arxiv_00833}
D.~Ghoshdastidar, M.~Gutzeit, A.~Carpentier, and U.~von Luxburg.
\newblock Two-sample hypothesis testing for inhomogeneous random graphs.
\newblock arXiv preprint (arXiv:1707.00833), 2017{\natexlab{a}}.

\bibitem[Ghoshdastidar et~al.(2017{\natexlab{b}})Ghoshdastidar, Gutzeit,
  Carpentier, and von Luxburg]{Ghoshdastidar_2017_conf_COLT}
D.~Ghoshdastidar, M.~Gutzeit, A.~Carpentier, and U.~von Luxburg.
\newblock Two-sample tests for large random graphs using network statistics.
\newblock In \emph{Conference on Learning Theory (COLT)}, 2017{\natexlab{b}}.

\bibitem[Ginestet et~al.(2014)Ginestet, Fournel, and
  Simmons]{Ginestet_2014_jour_FrontComputNeurosci}
C.~E. Ginestet, A.~P. Fournel, and A.~Simmons.
\newblock Statistical network analysis for functional {MRI}: Summary networks
  and group comparisons.
\newblock \emph{Frontiers in computational neuroscience}, 8\penalty0
  (51):\penalty0 10.3389/fncom.2014.00051, 2014.

\bibitem[Ginestet et~al.(2017)Ginestet, Li, Balachandran, Rosenberg, and
  Kolaczyk]{Ginestet_2017_jour_AOAS}
C.~E. Ginestet, J.~Li, P.~Balachandran, S.~Rosenberg, and E.~D. Kolaczyk.
\newblock Hypothesis testing for network data in functional neuroimaging.
\newblock \emph{The Annals of Applied Statistics}, 11\penalty0 (2):\penalty0
  725--750, 2017.

\bibitem[Goldreich et~al.(1998)Goldreich, Goldwasser, and
  Ron]{Goldreich_1998_jour_JACM}
O.~Goldreich, S.~Goldwasser, and D.~Ron.
\newblock Property testing and its connection to learning and approximation.
\newblock \emph{Journal of the ACM}, 45\penalty0 (4):\penalty0 653--750, 1998.

\bibitem[Gretton et~al.(2012)Gretton, Borgwardt, Rasch, Sch{\"o}lkopf, and
  Smola]{Gretton_2012_jour_JMLR}
A.~Gretton, K.~M. Borgwardt, M.~J. Rasch, B.~Sch{\"o}lkopf, and A.~Smola.
\newblock A kernel two-sample test.
\newblock \emph{Journal of Machine Learning Research}, 13:\penalty0 723--733,
  2012.

\bibitem[Hyduke et~al.(2013)Hyduke, Lewis, and
  Palsson]{Hyduke_2013_jour_MolBioSys}
D.~R. Hyduke, N.~E. Lewis, and B.~Palsson.
\newblock Analysis of omics data with genome-scale models of metabolism.
\newblock \emph{Molecular BioSystems}, 9\penalty0 (2):\penalty0 167--174, 2013.

\bibitem[Kondor and Pan(2016)]{Kondor_2016_conf_NIPS}
R.~Kondor and H.~Pan.
\newblock The multiscale {Laplacian} graph kernel.
\newblock In \emph{Advances in Neural Information Processing Systems (NIPS)},
  2016.

\bibitem[Landman et~al.(2011)Landman, Huang, Gifford, Vikram, Lim, Farrell,
  Bogovic, Hua, Chen, Jarso, Smith, Joel, Mori, Pekar, Barker, Prince, and {van
  Zijl}]{Landman_2011_jour_Neuroimage}
B.~A. Landman, A.~J. Huang, A.~Gifford, D.~S. Vikram, I.~A. Lim, J.~A. Farrell,
  J.~A. Bogovic, J.~Hua, M.~Chen, S.~Jarso, S.~A. Smith, S.~Joel, S.~Mori,
  J.~J. Pekar, P.~B. Barker, J.~L. Prince, and P.~C. {van Zijl}.
\newblock Multi-parametric neuroimaging reproducibility: A {3-T} resource
  study.
\newblock \emph{Neuroimage}, 54\penalty0 (4):\penalty0 2854--2866, 2011.

\bibitem[Lee and Yin(2014)]{Lee_2014_jour_DukeMathJ}
J.~O. Lee and J.~Yin.
\newblock A necessary and sufficient condition for edge universality of
  {Wigner} matrices.
\newblock \emph{Duke Mathematical Journal}, 163\penalty0 (1):\penalty0
  117--173, 2014.

\bibitem[Lei(2016)]{Lei_2016_jour_AnnStat}
J.~Lei.
\newblock A goodness-of-fit test for stochastic block models.
\newblock \emph{The Annals of Statistics}, 44\penalty0 (1):\penalty0 401--424,
  2016.

\bibitem[Leskovec and Krevl(2014)]{Leskovec_2014_data_snap}
J.~Leskovec and A.~Krevl.
\newblock {SNAP Datasets}: {Stanford} large network dataset collection.
\newblock \url{http://snap.stanford.edu/data}, 2014.

\bibitem[Leskovec et~al.(2005)Leskovec, Kleinberg, and
  Faloutsos]{Leskovec_2005_conf_KDD}
J.~Leskovec, J.~Kleinberg, and C.~Faloutsos.
\newblock Graphs over time: Densification laws, shrinking diameters and
  possible explanations.
\newblock In \emph{ACM SIGKDD International Conference on Knowledge Discovery
  and Data Mining}, 2005.

\bibitem[Lov{\'a}sz(2012)]{Lovasz_2012_book_AMS}
L.~Lov{\'a}sz.
\newblock \emph{Large networks and graph limits}.
\newblock American Mathematical Society, 2012.

\bibitem[Mukherjee et~al.(2017)Mukherjee, Sarkar, and
  Lin]{Mukherjee_2017_conf_NIPS}
S.~S. Mukherjee, P.~Sarkar, and L.~Lin.
\newblock On clustering network-valued data.
\newblock In \emph{Advances in Neural Information Processing Systems (NIPS)},
  2017.

\bibitem[Ng et~al.(2002)Ng, Jordan, and Weiss]{Ng_2002_conf_NIPS}
A.~Ng, M.~I. Jordan, and Y.~Weiss.
\newblock On spectral clustering: Analysis and an algorithm.
\newblock In \emph{Advances in Neural Information Processing Systems (NIPS)},
  2002.

\bibitem[Shervashidze et~al.(2011)Shervashidze, Schweitzer, {van Leeuwen},
  Mehlhorn, and Borgwardt]{Shervashidze_2011_jour_JMLR}
N.~Shervashidze, P.~Schweitzer, E.~J. {van Leeuwen}, K.~Mehlhorn, and K.~M.
  Borgwardt.
\newblock Weisfeiler-{Lehman} graph kernels.
\newblock \emph{Journal of Machine Learning Research}, 12:\penalty0 2539--2561,
  2011.

\bibitem[Tang et~al.(2016)Tang, Athreya, Sussman, Lyzinski, and
  Priebe]{Tang_2017_jour_JCompGraphStat}
M.~Tang, A.~Athreya, D.~L. Sussman, V.~Lyzinski, and C.~E. Priebe.
\newblock A semiparametric two-sample hypothesis testing problem for random
  graphs.
\newblock \emph{Journal of Computational and Graphical Statistics}, 26\penalty0
  (2):\penalty0 344--354, 2016.

\bibitem[Tang et~al.(2017)Tang, Athreya, Sussman, Lyzinski, and
  Priebe]{Tang_2017_jour_Bernoulli}
M.~Tang, A.~Athreya, D.~L. Sussman, V.~Lyzinski, and C.~E. Priebe.
\newblock A nonparametric two-sample hypothesis testing problem for random
  graphs.
\newblock \emph{Bernoulli}, 23:\penalty0 1599--1630, 2017.

\bibitem[Tracy and Widom(1996)]{Tracy_1996_jour_CommMathPhys}
C.~A. Tracy and H.~Widom.
\newblock On orthogonal and symplectic matrix ensembles.
\newblock \emph{Communications in Mathematical Physics}, 177:\penalty0
  727--754, 1996.

\bibitem[Yang and Leskovec(2013)]{Yang_2013_conf_WSDM}
J.~Yang and J.~Leskovec.
\newblock Overlapping community detection at scale: {A} nonnegative matrix
  factorization approach.
\newblock In \emph{Proceedings of the sixth ACM international conference on Web
  search and data mining (WSDM)}, pages 587--596, 2013.

\bibitem[Zhang et~al.(2009)Zhang, Li, Riggins, Zhan, Xuan, Zhang, Hoffman,
  Clarke, and Wang]{Zhang_2009_jour_Bioinformatics}
B.~Zhang, H.~Li, R.~B. Riggins, M.~Zhan, J.~Xuan, Z.~Zhang, E.~P. Hoffman,
  R.~Clarke, and Y.~Wang.
\newblock Differential dependency network analysis to identify
  condition-specific topological changes in biological networks.
\newblock \emph{Bioinformatics}, 25\penalty0 (4):\penalty0 526--532, 2009.

\end{thebibliography}

\newpage
\appendix
{
\large
\textbf{Appendix}
\vskip0.5ex
}

Here, we provide additional details such as proofs, description of tests, additional numerical results and discussions.
Section~\ref{sec_proof} provides proofs for the theorems stated in the paper along with a corollary of Theorem~\ref{thm_tw}.
Section~\ref{sec_algo} provides detailed descriptions of all tests considered in our implementations, both existing tests as well as proposed ones.
Section~\ref{sec_addexpt} provides additional numerical results, which we have referred to in the paper. 

\section{Proofs for results}
\label{sec_proof}

In this section, we present the proofs for Theorems~\ref{thm_normal} and~\ref{thm_tw}, which provide the theoretical foundations for the proposed tests \alg{Asymp-Normal} and \alg{Asymp-TW}, respectively.

\subsection{Proof of Theorem~\ref{thm_normal}}
For convenience, we assume $m$ is even. The extension to odd $m$ is straightforward.
We also write $P,Q$ instead of $P^{(n)},Q^{(n)}$ and define 
\begin{align*}
\widehat{\mu}_{ij} &=  \left( \sum\limits_{k\leq m/2} (A_{G_k})_{ij} -  (A_{H_k})_{ij} \right)\left( \sum\limits_{k> m/2} (A_{G_k})_{ij} -  (A_{H_k})_{ij} \right),
\\ \widehat{s}_{ij}^2 &=  \left( \sum\limits_{k\leq m/2} (A_{G_k})_{ij} +  (A_{H_k})_{ij} \right)\left( \sum\limits_{k> m/2} (A_{G_k})_{ij} +  (A_{H_k})_{ij} \right),
\\ \widehat\mu &= \sum_{i<j} \widehat\mu_{ij}, \qquad \text{ and } \qquad \widehat{s} = \sqrt{\sum_{i<j} \widehat{s}^2_{ij}}.
\end{align*}
Also let $\mu = \mathbb{E}[\widehat\mu] = \frac{m^2}{8}\Vert P-Q\Vert_F^2$, $s^2 = \mathbb{E}[\widehat{s}^2] = \frac{m^2}{8}\Vert P+Q\Vert_F^2$, and $\sigma^2 = \sum\limits_{i<j} \text{Var}(\widehat\mu_{ij})$.

Under the null hypothesis, that is $P=Q$, $\{\widehat\mu_{ij}: i<j\}$ are centred mutually independent random variables, and hence, 
due to the central limit theorem, we can claim that $\frac{\widehat\mu}{\sigma}$ converges to a standard normal random variable as $n\to\infty$.
The rate of convergence is given by the Berry-Esseen theorem~\citep{Berry_1941_jour_AMSTrans} as
\begin{align*}
\sup_x \left| F_{\widehat\mu/\sigma}(x) - \Phi(x)\right| \leq \frac{10}{\sigma^3} \sum_{i<j} \mathbb{E}\left[|\widehat\mu_{ij}|^3\right],
\end{align*}
where $F_{\widehat\mu/\sigma}(\cdot)$ is the distribution function for $\frac{\widehat\mu}{\sigma}$. 
Recall our assumption that the entries are bounded away from 1. Let $\max\limits_{ij} P_{ij} \leq 1-\delta$ for some $\delta>0$. 
Observe that $\widehat\mu_{ij}$ is product of two i.i.d. random variables, where each of them is a difference of two binomials. Hence, under $\mathcal{H}_0$, we can compute
\begin{align*}
\sigma^2 &= \sum_{i<j} \left(\frac{m}{2}2P_{ij}(1-P_{ij})\right)^2 \geq \frac{m^2\delta^2}{2}\Vert P\Vert_F^2,
\end{align*}
and by using the Cauchy-Schwarz inequality,
\begin{align*}
\mathbb{E}\left[|\widehat\mu_{ij}|^3\right] 
&\leq \sqrt{\mathbb{E}\left[\widehat\mu_{ij}^2\right]\mathbb{E}\left[\widehat\mu_{ij}^4\right]}
\\&= mP_{ij}(1-P_{ij})\left(mP_{ij}(1-P_{ij})^3 + \frac{m}{2}\left(\frac{m}{2}-1\right)4P_{ij}^2(1-P_{ij})^2\right)
\\&\leq m^2P_{ij}^2 + m^3P_{ij}^3 \leq 2m^3P_{ij}^2.
\end{align*}
Hence, the Berry-Esseen bound can be written as
\begin{align*}
\sup_x \left| F_{\widehat\mu/\sigma}(x) - \Phi(x)\right| \leq 20\sqrt{2}\frac{m^3\Vert P\Vert_F^2}{m^3\delta^3\Vert P\Vert_F^3} = o_n(1)
\end{align*}
since $\Vert P\Vert_F = \omega_n(1)$. We now compute the probability of type-I error in the following way:
\begin{align}
\mathbb{P}\big( T_{fro} \notin [-t_\alpha,t_\alpha] \big) 
&= \mathbb{P}\left(\frac{|\widehat\mu|}{\widehat{s}} > t_\alpha\right) 
\leq \mathbb{P}\left(\frac{|\widehat\mu|}{\sigma} > (1-\epsilon)t_\alpha\right)  + 
\mathbb{P}\left( \widehat{s}^2 < (1-\epsilon)^2\sigma^2\right)
\label{eqn_pf1}
\end{align}
for any $\epsilon\in(0,\frac12)$. Using the Berry-Esseen bound, we bound the first term as
\begin{align*}
\mathbb{P}\left(\frac{|\widehat\mu|}{\sigma} > (1-\epsilon)t_\alpha\right) 
&= 2\big(1-\Phi((1-\epsilon)t_\alpha)\big) + 2\left| F_{\widehat\mu/\sigma}((1-\epsilon)t_\alpha) - \Phi((1-\epsilon)t_\alpha)\right|
\\&= \alpha +  2\big(\Phi(t_\alpha)-\Phi((1-\epsilon)t_\alpha)\big) + o_n(1)
\\&\leq \alpha + \epsilon t_\alpha\sqrt{\frac2\pi}\exp\left(-\frac{t^2_\alpha}{8}\right) + o_n(1) 
\end{align*}
where we use $\epsilon\leq \frac12$ in the last step. 
Taking $\epsilon = \Vert P\Vert_F^{-1/2}$ leads to a bound $\alpha + o_n(1)$.

We now deal with the second term in~\eqref{eqn_pf1}.
Observe that $\sigma^2 \leq \frac{m^2}{2}\Vert P\Vert_F^2 \leq s^2$. Hence, we have
\begin{align*}
\mathbb{P}\left( \widehat{s}^2 < (1-\epsilon)^2\sigma^2\right)
&\leq \mathbb{P}\left( \widehat{s}^2 < (1-\epsilon)s^2\right)
\\&= \mathbb{P}\left( s^2 - \widehat{s}^2 > \epsilon s^2\right)
\leq \frac{\text{Var}(\widehat{s}^2)}{\epsilon^2 s^4}
\end{align*}
by the Chebyshev inequality.
We can compute the variance term for any $P,Q$ as
\begin{align}
\label{eqn_pf2}
&\text{Var}(\widehat{s}^2)
\\&= \sum_{i<j} \frac{m^2}{4}\left(P_{ij}(1-P_{ij}) + Q_{ij}(1-Q_{ij})\right)^2 + \frac{m^3}{4}(P_{ij}+Q_{ij})^2\left(P_{ij}(1-P_{ij}) + Q_{ij}(1-Q_{ij})\right)
\nonumber
\end{align}
In particular, under $\mathcal{H}_0$, $\text{Var}(\widehat{s}^2) \leq 2m^3\Vert P\Vert_F^2$.
Using this, the Chebyshev bound is smaller than $\frac{4}{m\epsilon^2\Vert P\Vert_F^2} = o_n(1)$ for $\epsilon = \Vert P\Vert_F^{-1/2}$.
Hence, we obtained the claimed type-I error bound.

For the type-II error rate, we consider the stated separation condition in the form $\frac{m\Vert P-Q\Vert_F^2}{\Vert P+Q\Vert_F} = \omega_n(1)$. 
We can bound the error probability as
\begin{align*}
\mathbb{P}\big( T_{fro} \in [-t_\alpha,t_\alpha] \big) 
\leq \mathbb{P}\left(\frac{|\widehat\mu|}{s} \leq 2t_\alpha\right)  + 
\mathbb{P}\left( \widehat{s}^2 \geq 4s^2\right).
\end{align*}
For the second term, we use the Chebyshev inequality as above to show that the probability is $o_n(1)$ since $\Vert P+Q\Vert_F = \omega_n(1)$. For the first term, observe that we have $\frac{\mu}{s} = \omega_n(1)$ under the separation condition, and hence for any fixed $\alpha$, we have $2t_\alpha \leq \frac{\mu}{2s}$ for large enough $n$. So,
\begin{align*}
\mathbb{P}\left(\frac{|\widehat\mu|}{s} \leq 2\tau_\alpha\right) 
&\leq \mathbb{P}\left(\frac{\widehat\mu}{s} \leq \frac{\mu}{2s}\right) 
\leq \frac{4\text{Var}(\widehat{\mu})}{\mu^2}\;.
\end{align*}
One can compute $\text{Var}(\widehat{\mu})$ similar to~\eqref{eqn_pf2} to obtain
\begin{align*}
\text{Var}(\widehat{\mu})
&\leq \sum_{i<j} \frac{m^2}{4}(P_{ij}+Q_{ij})^2 + \frac{m^3}{4}(P_{ij}-Q_{ij})^2(P_{ij}+Q_{ij})
\\&\leq \frac{m^2}{8}\Vert P+Q\Vert_F^2 + \frac{m^3}{8}\Vert P-Q\Vert_F^2\Vert P+Q\Vert_F,
\end{align*}
where the second inequality follows from use of the Cauchy-Schwarz inequality followed by the observation that $\ell_4$-norm is smaller than $\ell_2$-norm. Hence, the error probability is bounded as
\begin{align*}
\mathbb{P}\big( T_{fro} \in [-t_\alpha,t_\alpha] \big) 
\leq 32\frac{m^2\Vert P+Q\Vert_F^2 + m^3\Vert P-Q\Vert_F^2\Vert P+Q\Vert_F}{m^4\Vert P-Q\Vert_F^4} + o_n(1) = o_n(1)
\end{align*}
under the assumed separation. Hence, the claim.

\subsection{Proof of Theorem~\ref{thm_tw}}

We first derive the asymptotic distribution under the null hypothesis. This part is similar to the proof of Lemma A.1 in~\citet{Lei_2016_jour_AnnStat}.
Observe that under $\mathcal{H}_0$, $C$ in~\eqref{eqn_diffmatrix} is a symmetric random matrix, whose entries above the diagonal are independent with mean zero and variance $\frac{1}{n-1}$.
Now, let $D$ be a symmetric random matrix with zero diagonal, whose entries above the diagonal are i.i.d. normal with mean zero and variance $\frac{1}{n-1}$.
Due to the results of \citet{Erdos_2012_jour_AdvMath}, we know that $\lambda_1(C)$ and $\lambda_1(D)$ have the same limiting distribution. 
\citet{Lee_2014_jour_DukeMathJ} show that $n^{2/3}(\lambda_1(D)-2)\to TW_1$ as $n\to\infty$, and hence the same conclusion holds for $n^{2/3}(\lambda_1(C)-2)$.
The corresponding result for $-\lambda_n(C)$ can be proved by considering the matrix $-C$.
Based on this asymptotic result, we have
\begin{align*}
&\mathbb{P}\left( n^{2/3}(\lambda_1(C)-2) > \tau_\alpha \right) = \frac\alpha2 + o_n(1), \text{ and}
\\
&\mathbb{P}\left( n^{2/3}(-\lambda_n(C)-2) > \tau_\alpha \right) = \frac\alpha2 + o_n(1),
\end{align*}
where $\tau_\alpha$ is the $\frac\alpha2$ upper quantile of the $TW_1$ distribution.
Since, $\Vert C\Vert_2 = \max\left\{\lambda_1(C),-\lambda_n(C)\right\}$, an union bound leads to the stated conclusion under the null hypothesis.

Under the alternative hypothesis, one can see that $\mathbb{E}[C]$ is a re-scaled version of $P-Q$ with each entry being scaled by normalising term of $\sqrt{(n-1)(P_{ij}(1-P_{ij}) + Q_{ij}(1-Q_{ij}))}$ (we drop the superscript $n$ for convenience).
Under the stated separation condition on $\left\Vert \mathbb{E}[C] \right\Vert_2$, it is easy to see that $n^{2/3}(\Vert C\Vert_2-2)\to\infty$ with high probability.
So, the probability of the test statistic being smaller than $\tau_\alpha$ is $o_n(1)$.
To be precise, we decompose $C$ as $C = \mathbb{E}[C] + (C-\mathbb{E}[C])$, and
using Weyl's inequality, we can write
\begin{align*}
\Vert C\Vert_2 &\geq \Vert \mathbb{E}[C]\Vert_2 - \Vert C-\mathbb{E}[C]\Vert_2
\geq \Vert \mathbb{E}[C]\Vert_2 - \left(2 + n^{-2/3}\tau_\beta\right) 
\end{align*}
with probability at most $\beta + o_n(1)$.
The second inequality follows by noting that $(C-\mathbb{E}[C])$ is a mean zero matrix whose spectral norm can be bounded using the arguments stated under the null hypothesis.
Hence, $n^{2/3}(\Vert C\Vert_2 - 2) \geq n^{2/3} (\Vert \mathbb{E}[C]\Vert_2 - 4)- \tau_\beta$ with probability $\beta + o_n(1)$. 
We set $\tau_\beta = n^{2/3}(\Vert \mathbb{E}[C]\Vert_2 -4)- \tau_\alpha$, and observe that $\tau_\beta = \omega_n(1)$, that is $\beta = o_n(1)$, if $\Vert \mathbb{E}[C]\Vert_2 \geq 4 + \omega_n(n^{-2/3})$. 

\subsection{Theorem~\ref{thm_tw} for stochastic block models}
\label{sec_thm_tw_blockmodel}

We state the following corollary, which provides an understanding of the condition on $\mathbb{E}[C]$ in Theorem~\ref{thm_tw} under a block model assumption.

\begin{cor}
Assume that $P^{(n)},Q^{(n)}$ correspond to stochastic block models with at most $r_n$ communities, and
let $\rho_n = \max\limits_{ij} \left\{P_{ij}^{(n)},Q_{ij}^{(n)}\right\}$ .
If $\left\Vert P^{(n)} - Q^{(n)} \right\Vert_F^2 = \omega_n\left(nr_n^2\rho_n\right)$, then  
\begin{align}
\mathbb{P}\left( n^{2/3}(\Vert C\Vert_2-2) \leq \tau_\alpha \right) = o_n(1).
\end{align}
\end{cor}

One can observe that if $r_n$ is bounded by a constant and all entries of $P^{(n)},Q^{(n)}$ are of the same order (same as $\rho_n$), then the above separation condition is similar to the one stated in Theorem~\ref{thm_normal}.

\begin{proof}
The claim would follow if we show that under the stated separation, the condition on $\mathbb{E}[C]$ used in Theorem~\ref{thm_tw} holds.
In fact, we show that in the present case, $\Vert \mathbb{E}[C]\Vert_2 = \omega_n(1)$.
For convenience, we simply write $P,Q$ and define $R_{ij} = \sqrt{(n-1)(P_{ij}(1-P_{ij}) + Q_{ij}(1-Q_{ij}))} \leq \sqrt{2n\rho_n}$.
Note that
\begin{align*}
\mathbb{E}[C_{ij}] = \frac{P_{ij} - Q_{ij}}{R_{ij}} \;,
\end{align*}
and hence, $\mathbb{E}[C]$ has a block structure with at most $r_n^2$ blocks (ignoring that the diagonal is zero).
Thus, there is a diagonal matrix $\Lambda$ such that $\Lambda+\mathbb{E}[C]$ has rank at most $r_n^2$.
Note that the diagonal entries of $\Lambda$ are same as the diagonal blocks of $C$, and so, $\Vert \Lambda\Vert_2 \leq \max\limits_{ij} \frac{|P_{ij} - Q_{ij}|}{R_{ij}} \leq 2\sqrt{\frac{\rho_n}{(n-1)(1-\rho_n)}} = o_n(1)$
assuming that $\rho_n$ is bounded away from 1.
Hence, we can write
\begin{align*}
\Vert \mathbb{E}[C] \Vert_2 \geq \Vert \Lambda+ \mathbb{E}[C] \Vert_2 - \Vert \Lambda\Vert_2
&\geq \frac{1}{r_n} \Vert \Lambda+\mathbb{E}[C] \Vert_F - o_n(1)
\\&\geq \frac{1}{r_n} \Vert \mathbb{E}[C] \Vert_F - o_n(1) \geq \frac{\Vert P-Q\Vert_F}{r_n\sqrt{2n\rho_n}} - o_n(1),
\end{align*}
which is $\omega_n(1)$ under the stated  condition.
For the second inequality, we use the relation between spectral and Frobenius norms of a matrix with rank $r_n^2$.
Finally, Theorem~\ref{thm_tw} leads to the result.
\end{proof}

\section{Detailed description of tests}
\label{sec_algo}

In this section, we describe all the tests discussed in this paper.
First, we provide description of the asymptotic tests, which include the tests \alg{Asymp-Normal} and \alg{Asymp-TW} proposed in this paper, as well as the large-sample test \alg{Asymp-Chi2}.
We next describe the bootstrapped tests \alg{Boot-Spectral} and \alg{Boot-Frobenius}, which are based on approximating the null distribution by randomly permuting the group assignments of the graphs.
\citet{Tang_2017_jour_JCompGraphStat} provide an algorithmic description of \alg{Boot-ASE}. 
For completeness, we include this description along with that of \alg{Boot-EPA}, which also generates bootstrap samples based on a low rank approximation of population adjacency.
Throughout this section, we refer to the null hypothesis $\mathcal{H}_0$ as the hypotheses that both graphs (or graph populations) have the same population adjacency. 

\subsection{Asymptotic tests}

We first describe the \alg{Asymp-Normal} test below. 
In addition to accepting or rejecting the null hypothesis, we also present how to compute the \emph{p-value}, which is defined as the probability that the null hypothesis is true.
This is often useful to quantify the amount of dissimilarity between two populations.
We use the standard rule of rejecting the null hypothesis when $\pv$ is less than the prescribed significance level $\alpha$.
Note that in \alg{Asymp-Normal}, the $\pv$  involves a factor of 2 to take into account both the upper and the lower tail probabilities.

\begin{varalgorithm}{\alg{Asymp-Normal}}
\caption {~}
\label{test_normal}
\begin{algorithmic}[1]
 \REQUIRE Graphs $G_1,\ldots,G_m$ and $H_1,\ldots,H_m$ defined on a common vertex set $V$, where $m>1$;
 Significance level $\alpha$.\\
 \STATE Compute $T_{fro}$ as shown in~\eqref{eqn_stat_fro}.\\
 \STATE $\pv=2\big(1-\Phi(-|T_{fro}|)\big)$, where $\Phi$ is the standard normal distribution function.\\
 \ENSURE Reject the null hypothesis if $\pv\leq \alpha$.\\
\end{algorithmic}
\end{varalgorithm}

The  \alg{Asymp-Chi2} test is listed below.
For convenience,  we write $T_{\chi^2} = \sum\limits_{i<j} \frac{\tilde\mu^2_{ij}}{\tilde\sigma^2_{ij}}$, where $\tilde\mu^2_{ij}$ and $\tilde\sigma^2_{ij}$ denote the numerator and denominator of each term in the summation~\eqref{eqn_stat_chi2}.
This notation corresponds to the fact that $\tilde\mu_{ij}$ is the sample mean difference for entry $(i,j)$, and $\tilde\sigma^2_{ij}$ is an estimate of the variance of $\tilde\mu_{ij}$.
We note that for sparse graphs and small $m$, the summation in~\eqref{eqn_stat_chi2} may have terms of the form $\frac00$. Hence, we sum only over the set of edges in $\mathcal{C}$ defined below.

\begin{varalgorithm}{\alg{Asymp-Chi2}}
\caption {~}
\label{test_chi2}
\begin{algorithmic}[1]
 \REQUIRE Graphs $G_1,\ldots,G_m$ and $H_1,\ldots,H_m$, where $m>1$; Significance level $\alpha$.\\
 \STATE Let $\mathcal{C} = \left\{(i,j): i<j,~ \tilde\mu_{ij}\neq0~ \text{or} ~\tilde\sigma_{ij}\neq0\right\}$, where $\tilde\mu_{ij}$, $\tilde\sigma_{ij}$ are defined above.\\
 \STATE Compute $T_{\chi^2}$ similar to~\eqref{eqn_stat_chi2}, but sum only over  $(i,j)\in\mathcal{C}$.\\
 \STATE $\pv=1- F_{\chi^2}\left(T_{\chi^2},\frac{n(n-1)}{2}\right)$, where $F_{\chi^2}(\cdot,\nu)$ is the $\chi^2$-distribution function with degree of freedom $\nu$.\\
 \ENSURE Reject the null hypothesis if $\pv\leq \alpha$.\\
\end{algorithmic}
\end{varalgorithm}

We now described \alg{Asymp-TW}, which is the proposed asymptotic test for testing between two given graphs $G$ and $H$ (that is, $m=1$).
As noted in the main paper, this test uses a block model approximation to compute the matrices $\widetilde{P},\widetilde{Q}$.
In the following description, we assume that a partition of $V$ into $V_1,\ldots,V_r$ is provided as input to the test.
For simplicity, we assume that the same partitioning is used for both graphs, but this is not a necessity.
In our implementations, we use normalised spectral clustering~\citep{Ng_2002_conf_NIPS} to compute the partition from the average of the two adjacency matrices.
A minor difference here is that we use the dominant singular vectors  of the normalised adjacency instead of the dominant eigenvectors.
This modification is made since the networks could be either homophilic (communities are highly connected) or heterophilic (inter-community links are more frequent as in a bi-partite graph).
We also provide an option to externally provide the communities. We use this feature for the real data from Stanford network collection, where we pre-compute the community structure using BigClam~\citep{Yang_2013_conf_WSDM}.
From the test statistic $T_{TW}$, we compute the $\pv$ by using available table of distribution function for Tracy-Widom law.\footnote{A table, based on \citet{Bornemann_2010_jour_MarkovProc} and containing values for $F_{TW_1}(\cdot)\leq 0.9998$, was obtained from \newline \url{http://www.wisdom.weizmann.ac.il/ ~nadler/Wishart_Ratio_Trace/TW_ratio.html}.} 
The factor of 2 is due to the fact that only the extreme eigenvalues are known to follow the $TW_1$ distribution, and hence, we need union bound for $\Vert\widetilde{C}\Vert_2= \max\left\{\lambda_1(\widetilde{C}),-\lambda_n(\widetilde{C})\right\}$. 

\begin{varalgorithm}{\alg{Asymp-TW}}
\caption {~}
\label{test_tw}
\begin{algorithmic}[1]
 \REQUIRE Graphs $G,H$ defined on vertex set $V$; Partition of $V$ into $V_1,\ldots,V_r$; Significance level $\alpha$.\\
 \FORALL{$V_k$}
 \FORALL{$i,j\in V_k, i\neq j$}
 \STATE Let $\widetilde{P}_{ij} = \frac{2}{|V_k|(|V_k|-1)} \sum\limits_{i',j'\in V_k: i'<j'} (A_G)_{ij}$ and $\widetilde{Q}_{ij} = \frac{2}{|V_k|(|V_k|-1)} \sum\limits_{i',j'\in V_k: i'<j'} (A_H)_{ij}$.\\
 \ENDFOR
 \ENDFOR
 \FORALL{$V_k,V_l, k\neq l$}
 \FORALL{$i\in V_k, j\in V_l$}
 \STATE Compute $\widetilde{P}_{ij} = \frac{1}{|V_k||V_l|} \sum\limits_{i'\in V_k, j'\in V_l} (A_G)_{ij}$ and $\widetilde{Q}_{ij} = \frac{1}{|V_k||V_l|} \sum\limits_{i'\in V_k, j'\in V_l} (A_H)_{ij}$.\\
 \ENDFOR
 \ENDFOR
 \STATE Compute $\widetilde{C}$ and $T_{TW}$ as in~\eqref{eqn_stat_tw}.\\
 \STATE $\pv=2\big(1- F_{TW_1}\left(T_{TW}\right)\big)$, where $F_{TW_1}$ is the distribution function for Tracy-Widom law.\\
 \ENSURE Reject the null hypothesis if $\pv\leq \alpha$.\\
\end{algorithmic}
\end{varalgorithm}

\subsection{Bootstrap tests}

We begin with the description of \alg{Boot-Spectral} and \alg{Boot-Frobenius}.
We present both tests together since they follow the same bootstrapping procedure, and only differ in terms of the test statistic.
The differences of \alg{Boot-Frobenius} from \alg{Boot-Spectral} are noted in parentheses.

\begin{varalgorithm}{\alg{Boot-Spectral} (or \alg{Boot-Frobenius})}
\caption {~}
\label{test_boot_permute}
\begin{algorithmic}[1]
 \REQUIRE Graphs $G_1,\ldots,G_m$ and $H_1,\ldots,H_m$, where $m>1$;
 Significance level $\alpha$; Number of bootstraps $b$.\\
 \STATE Let $T = T_{spec}$ as computed in~\eqref{eqn_stat_spec} (or $T=T_{fro}$ in~\eqref{eqn_stat_fro}).\\
 \FOR{$i=1$ \TO $b$} 
 \STATE Randomly split $\{G_1,\ldots,G_m,H_1,\ldots,H_m\}$ into two populations of equal size.\\
 \STATE Let $T_i$ be the spectral norm statistic~\eqref{eqn_stat_spec} for this split (or Frobenius norm statistic~\eqref{eqn_stat_fro}).\\
 \ENDFOR
 \STATE $\pv=\frac1b\left(\big|\{i:T_i \geq T\}\big|+0.5\right)$, where 0.5 is added for continuity correction.\\
 \ENSURE Reject the null hypothesis if $\pv\leq \alpha$.\\
\end{algorithmic}
\end{varalgorithm}

Finally, we present the tests \alg{Boot-ASE} and \alg{Boot-EPA} based on adjacency spectral embedding (ASE) and estimated population adjacency (EPA), respectively. 
The differences of \alg{Boot-EPA} from \alg{Boot-ASE} are noted in parentheses.
Note that these tests compute two approximations of the null distribution --- one based on pairs of graphs generated from $\widehat{P}$, and other based on graph pairs generated from $\widehat{Q}$.
The $\pv$ is finally computed to ensure that the null is rejected only when the test statistic is in the upper $\alpha$-quantile for both approximate distributions.

\begin{varalgorithm}{\alg{Boot-ASE} (or \alg{Boot-EPA})}
\caption {~}
\label{test_boot_lowrank}
\begin{algorithmic}[1]
 \REQUIRE Graphs $G$ and $H$; Significance level $\alpha$; Number of bootstraps $b$.\\
 \STATE Let $X_G$ and $\widehat{P}$ be the ASE and EPA for graph $G$, respectively (as described in Section~\ref{sec_m_one}).\\
 \STATE Let $X_H$ and $\widehat{Q}$ be the ASE and EPA for graph $H$, respectively.\\
 \STATE Compute $T = T_{ASE}$ as in~\eqref{eqn_stat_ase} (or $T=T_{EPA}$ in~\eqref{eqn_stat_epa}).\\
 \FOR{$i=1$ \TO $b$} 
 \STATE Generate $G_1,G_2\siid \ier(\widehat{P})$.\\
 \STATE Let $T_i$ be the ASE statistic~\eqref{eqn_stat_ase} between $G_1,G_2$ (or EPA statistic~\eqref{eqn_stat_epa}).\\
 \ENDFOR
 \STATE Compute $p =\frac1b\left(\big|\{i:T_i \geq T\}\big|+0.5\right)$, where 0.5 is added for continuity correction.\\
 \FOR{$i=1$ \TO $b$} 
 \STATE Generate $H_1,H_2\siid \ier(\widehat{Q})$.\\
 \STATE Let $T'_i$ be the ASE statistic~\eqref{eqn_stat_ase} between $H_1,H_2$ (or EPA statistic~\eqref{eqn_stat_epa}).\\
 \ENDFOR
 \STATE Compute $p' =\frac1b\left(\big|\{i:T'_i \geq T\}\big|+0.5\right)$.\\
 \STATE $\pv = \max\{p,p'\}$.\\
 \ENSURE Reject the null hypothesis if $\pv\leq \alpha$.\\
\end{algorithmic}
\end{varalgorithm}

\section{Additional numerical results and discussions}
\label{sec_addexpt}

Here, we provide additional results along with further details for the experiments with real data.

\subsection{Further simulations for random graphs}

In this section, we present the figures related to experiments on block models, which we have referred to in the main paper.
We have earlier noted that \alg{Asymp-Chi2} has an erratic behaviour for small $m$. 
This is not surprising since the variance estimates used in~\eqref{eqn_stat_chi2} are not reliable for small $m$, particularly when the graphs are sparse.
We demonstrate this effect even for slightly larger $m$ by comparing \alg{Asymp-Chi2} and \alg{Asymp-Normal} for $m\in\{10,20,50,100,200\}$. The graph sizes are kept relatively small $n\in\{50,100,150,200\}$. 
The models are same as the ones used in the experiment of Figure~\ref{fig_1a}.

\begin{figure}[ht]
\centering
\begin{tabular}{cc@{}l}
\multicolumn{2}{c}{ } & \hskip6ex Under null hypothesis  \hskip7ex Under alternative hypothesis \\
\rotatebox{90}{\small\hskip8ex \alg{Asymp-Normal} \hskip13ex \alg{Asymp-Chi2}} & 
\rotatebox{90}{\hskip13ex Test power (null rejection rate)} & 
\includegraphics[width=0.8\textwidth]{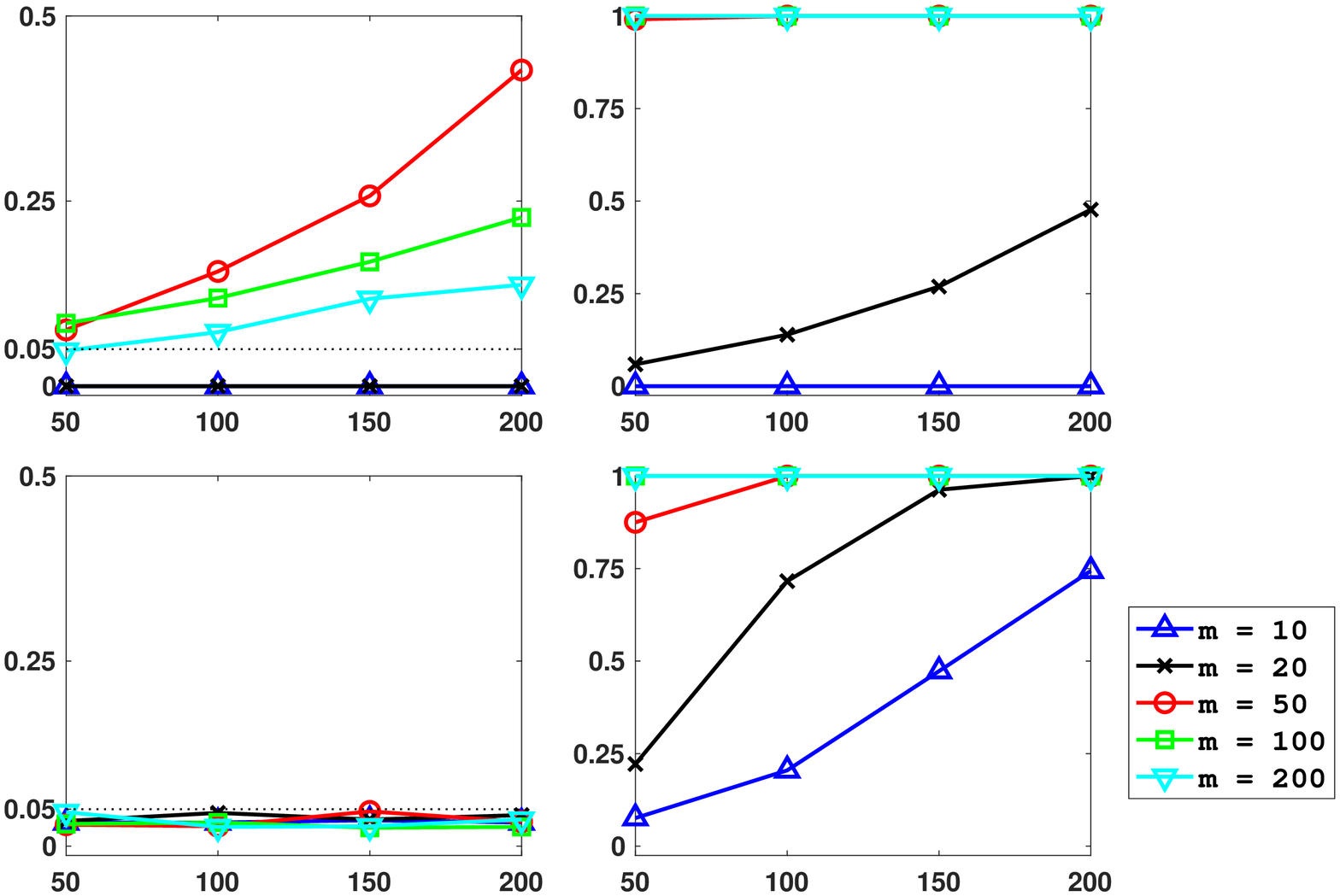}\\
& \multicolumn{2}{l}{\hskip22ex Number of vertices $n$}\\
\end{tabular}
\caption{Power of the asymptotic tests for different values of graph size $n$ and population size $m$. Each row corresponds to a particular test.}
\label{fig_1c}
\end{figure}

The result, plotted in Figure~\ref{fig_1c}, reveals the undesirable behaviour of \alg{Asymp-Chi2} as the test always has zero rejection rate for $m=10$. 
For $m\geq50$, the test power under alternative hypothesis is 1, but the rejection under null increases with $n$.
In particular, rejection rate under null is less than significance level only for $m=200$ and $n=50$.
Thus, \alg{Asymp-Chi2} is reliable only for $m\gg n$.
On the other hand, both Figures~\ref{fig_1a} and~\ref{fig_1c} confirm our theoretical observation that the behaviour of \alg{Asymp-Normal} under $\mathcal{H}_0$ does not change with $m$, while its power under $\mathcal{H}_1$ improves for large $m$.

\begin{figure}[t]
\centering
\begin{tabular}{cc@{}l}
\multicolumn{2}{c}{ } & \hskip6ex Under null hypothesis  \hskip7ex Under alternative hypothesis \\
\rotatebox{90}{\small\hskip7ex \alg{Asymp-Normal} \hskip12ex \alg{Boot-Frobenius} \hskip12ex \alg{Boot-Spectral}} & 
\rotatebox{90}{\hskip24ex Test power (null rejection rate)} & 
\includegraphics[width=0.8\textwidth]{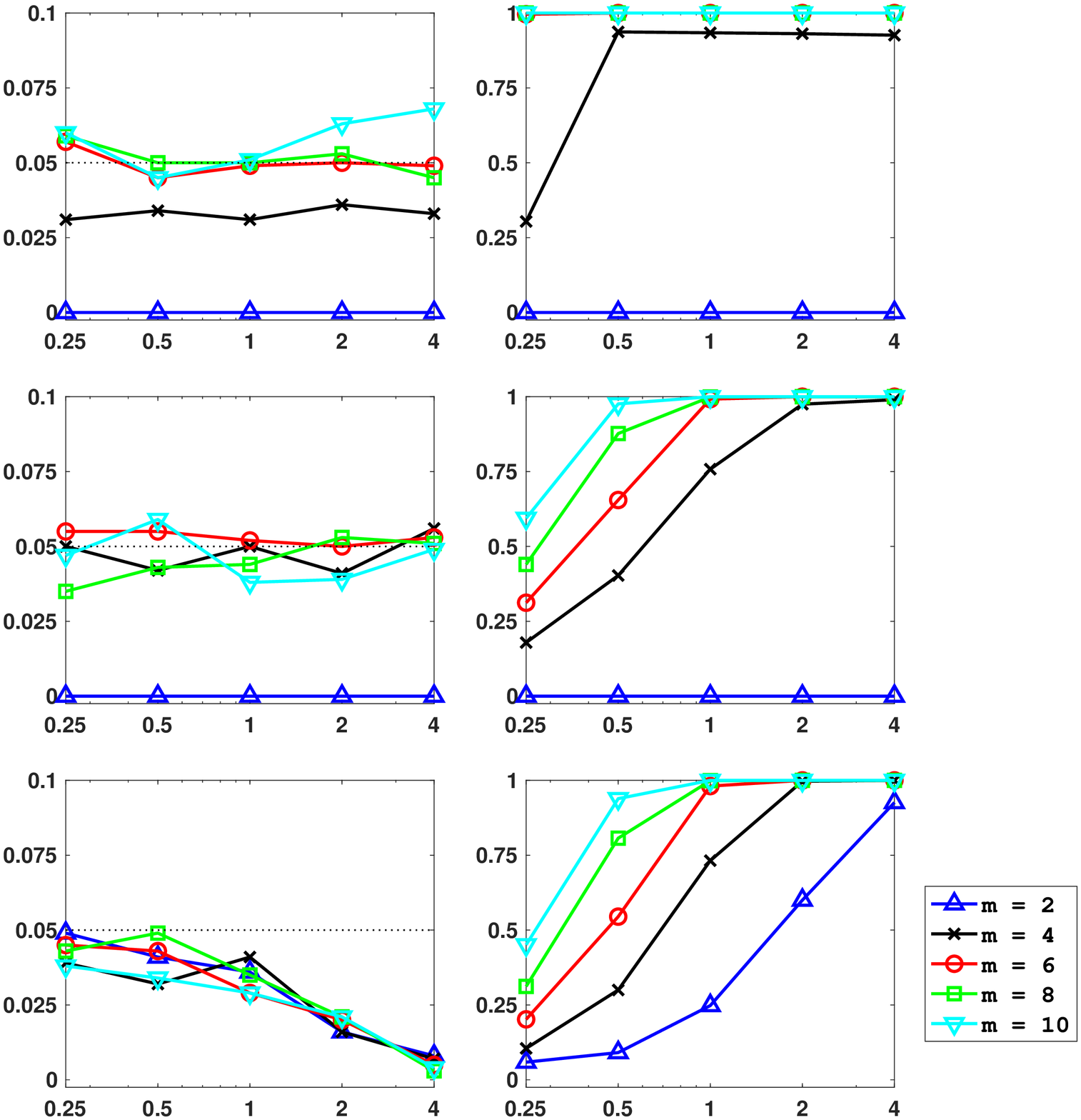}\\
& \multicolumn{2}{l}{\hskip25ex Density of graph $\rho$}\\
\end{tabular}
\caption{Power of different tests for varying levels of sparsity $\rho$ (larger $\rho$ implies denser graphs), and for different values of population size $m$. Each row corresponds to a particular test.}
\label{fig_1b}
\end{figure}

Figure~\ref{fig_1b} corresponds to our study related to varying levels of graph sparsity.
In this case, the models for $P^{(n)}$ and $Q^{(n)}$ are stochastic block models with same two communities. For $P^{(n)}$, within-class edge probability is $\rho p$ and across-class probability is $\rho q$.
We define $Q^{(n)}$ such that the within-class edge probability is $\rho(p+\epsilon)$.
In Figure~\ref{fig_1b}, we fix $n=500$ and show the rejection rates of the tests for varying sample size $m$ and density $\rho$.
The key conclusions are given in the main paper. Additionally, we note the effect of normal dominance in case of \alg{Asymp-Normal}.
Recall that $T_{fro}$ does not converge to the normal distribution, but it is dominated by a standard normal random variable. Thus, our threshold for rejection is actually higher than the $\frac\alpha2$-upper quantile of true asymptotic distribution of $T_{fro}$. This effect is pronounced for dense graphs, where the rejection rate under null  is much smaller than the pre-fixed 5\% level.

\begin{figure}[t]
\centering
\begin{tabular}{cc@{}l}
\multicolumn{2}{c}{ } & \hskip5ex Under null hypothesis  \hskip6ex Under alternative hypothesis \\
\rotatebox{90}{\hskip8ex \alg{Asymp-TW} \hskip13ex \alg{Boot-EPA} \hskip15ex \alg{Boot-ASE}} & 
\rotatebox{90}{\hskip24ex Test power (null rejection rate)} & 
\includegraphics[width=0.8\textwidth]{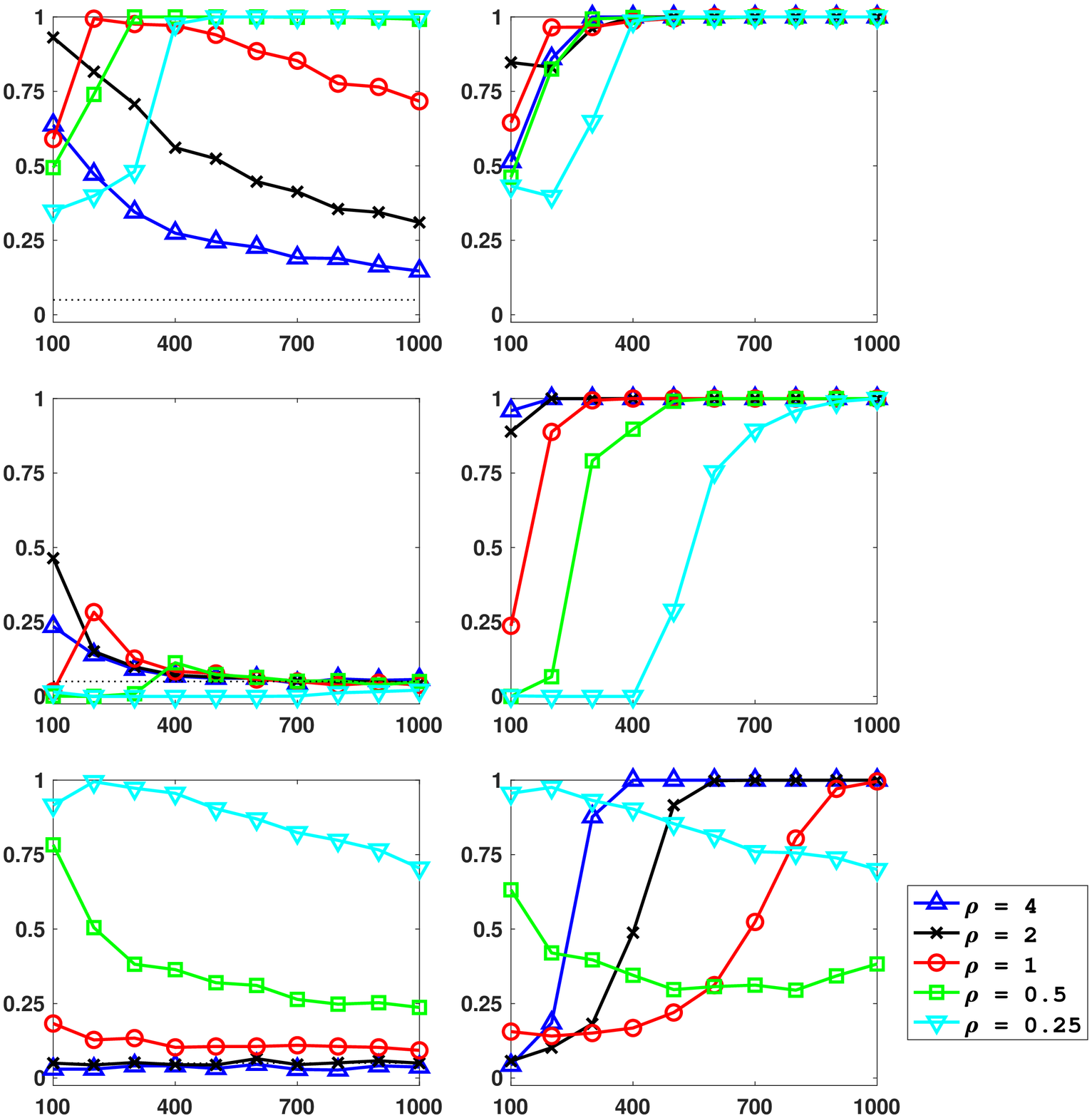}\\
& \multicolumn{2}{l}{\hskip22ex Number of vertices $n$}\\
\end{tabular}
\caption{Power of different tests with increase number of vertices $n$, and for different levels of sparsity $\rho$. Each row corresponds to a particular test.}
\label{fig_2b}
\end{figure}

We present a similar study on the effect of sparsity in the case $m=1$. The results in Figure~\ref{fig_2b} are based on the above setup, where we have $m=1$ and vary the the graph size $n$ and the density parameter $\rho$.
In this experiment, we use the true rank $r=2$. This provides an advantage to the bootstrap tests since we observe in Figure~\ref{fig_2a} that these tests fail when approximation based on a different rank is used.
We note that \alg{Boot-ASE} has a high rejection rate under both $\mathcal{H}_0$ and $\mathcal{H}_1$.
The rejection rate under $\mathcal{H}_0$ is smaller for dense graphs, but still above the desired 5\% level.
For sparse graphs $\rho<1$, this test is not reliable.
On the other hand, \alg{Boot-EPA} performs quite well for both sparse and dense graphs although it uses the same bootstrapping principle. Hence, we may conclude that the test statistic $T_{EPA}$, which was previously not used in the testing literature, is a more useful test statistic.
The asymptotic test \alg{Asymp-TW} works well for dense graphs $\rho\geq1$, but is not reliable in the sparse regime. There can be two potential reasons for this: (i) the approximation of normalisation terms in~\eqref{eqn_diffmatrix} using $\widetilde{P}$ and $\widetilde{Q}$ is poor in the sparse regime; or (ii) the use of standard spectral clustering for community detection fails in sparse graphs.
We believe that the latter reason is more probable since, in a later experiment with sparse real networks, we observe desirable performance from \alg{Asymp-TW} when the community detection is done using BigClam~\citep{Yang_2013_conf_WSDM}.

\subsection{Experiments with EEG recordings of epileptic seizure}
\label{sec_seizure}
\renewcommand{\arraystretch}{1.35}

In this section, we describe our experiments with networks constructed from EEG recordings of patients with epileptic seizure~\citep{Andrzejak_2001_jour_PhysRevE}.
We obtained the data from~\citet{Dua_2017_data_uci}, where each EEG recording is divided into several one-second snapshots containing 178 time points $(n=178)$.
There are a total of 11500 snapshots available that are classified into five groups:
\\{\bf Group-1:}
Recording of seizure activity;
\\{\bf Group-2:}
Recording of an area with tumour;
\\{\bf Group-3:}
Recording of a healthy brain area;
\\{\bf Group-4:}
Recording of patient with eyes open;
\\{\bf Group-5:}
Recording of patient with eyes closed.

\begin{table}[t]
\centering
\caption{Power of \alg{Asymp-Normal} for EEG correlation networks.}
\label{tab_seiz_pow2}
\begin{tabular}{|c|cc|cc|cc|cc|cc|}
\hline
	&	{\bf G1.1}	&	{\bf G1.2}	&	{\bf G2.1}	&	{\bf G2.2}	&	{\bf G3.1}	&	{\bf G3.2}	&	{\bf G4.1}	&	{\bf G4.2}	&	{\bf G5.1}	&	{\bf G5.2}	\\
\hline
{\bf G1.1}	&	0	&	0.011	&	1	&	1	&	1	&	1	&	1	&	1	&	1	&	1	\\
{\bf G1.2}	&	0.011	&	0	&	1	&	1	&	1	&	1	&	1	&	1	&	1	&	1	\\
\hline
{\bf G2.1}	&	1	&	1	&	0	&	0.003	&	0.009	&	0.008	&	1	&	1	&	1	&	1	\\
{\bf G2.2}	&	1	&	1	&	0.003	&	0	&	0.009	&	0.005	&	1	&	1	&	1	&	1	\\
\hline
{\bf G3.1}	&	1	&	1	&	0.009	&	0.009	&	0	&	0	&	1	&	1	&	1	&	1	\\
{\bf G3.2}	&	1	&	1	&	0.008	&	0.005	&	0	&	0	&	1	&	1	&	1	&	1	\\
\hline
{\bf G4.1}	&	1	&	1	&	1	&	1	&	1	&	1	&	0	&	0	&	1	&	1	\\
{\bf G4.2}	&	1	&	1	&	1	&	1	&	1	&	1	&	0	&	0	&	1	&	1	\\
\hline
{\bf G5.1}	&	1	&	1	&	1	&	1	&	1	&	1	&	1	&	1	&	0	&	0.010	\\
{\bf G5.2}	&	1	&	1	&	1	&	1	&	1	&	1	&	1	&	1	&	0.010	&	0	\\
\hline
\end{tabular}
\end{table}

\begin{table}[b]
\centering
\caption{Power of \alg{Asymp-TW} for EEG correlation networks.}
\label{tab_seiz_pow1}
\begin{tabular}{|c|cc|cc|cc|cc|cc|}
\hline
	&	{\bf G1.1}	&	{\bf G1.2}	&	{\bf G2.1}	&	{\bf G2.2}	&	{\bf G3.1}	&	{\bf G3.2}	&	{\bf G4.1}	&	{\bf G4.2}	&	{\bf G5.1}	&	{\bf G5.2}	\\
\hline
{\bf G1.1}	&	0	&	1	&	1	&	1	&	1	&	1	&	1	&	1	&	1	&	1	\\
{\bf G1.2}	&	1	&	0	&	1	&	1	&	1	&	1	&	1	&	1	&	1	&	1	\\
\hline
{\bf G2.1}	&	1	&	1	&	0	&	0.002	&	0	&	0.001	&	1	&	1	&	0.243	&	0.260	\\
{\bf G2.2}	&	1	&	1	&	0.002	&	0	&	0	&	0.001	&	1	&	1	&	0.247	&	0.251	\\
\hline
{\bf G3.1}	&	1	&	1	&	0	&	0	&	0	&	0	&	1	&	1	&	0.234	&	0.245	\\
{\bf G3.2}	&	1	&	1	&	0.001	&	0.001	&	0	&	0	&	1	&	1	&	0.243	&	0.258	\\
\hline
{\bf G4.1}	&	1	&	1	&	1	&	1	&	1	&	1	&	0	&	0.029	&	0.699	&	0.664	\\
{\bf G4.2}	&	1	&	1	&	1	&	1	&	1	&	1	&	0.029	&	0	&	0.647	&	0.667	\\
\hline
{\bf G5.1}	&	1	&	1	&	0.243	&	0.247	&	0.234	&	0.243	&	0.699	&	0.647	&	0	&	0.049	\\
{\bf G5.2}	&	1	&	1	&	0.260	&	0.251	&	0.245	&	0.258	&	0.664	&	0.667	&	0.049	&	0	\\
\hline
\end{tabular}
\end{table}

In our experiments, we construct networks by thresholding the autocorrelation matrices of the EEG snapshots.
The reason for considering such networks is due to their ubiquity in bioinformatics and neuroscience, where most networks are typically derived from correlations or covariances.
Moreover, through this setup, we also establish that though the proposed tests are theoretically analysed for edge-independent graphs, they can also be used for other types of networks.

We randomly split each class into four parts of equal size, and compute autocorrelation matrices from the snapshots in each part.
Unweighted graphs are obtained by retaining only the largest 10\% of correlations  (total of 20 graphs).
For \alg{Asymp-Normal} test, two graphs are needed for each population. Hence, for each class-$i$, we create two sub-groups {\bf G$i$.1} and {\bf G$i$.2}, each with two networks.
We subsequently test between every pair of the 10 sub-groups --- {\bf G$i$.1} vs. {\bf G$i$.2} is an instance of null hypothesis while every other pair is an instance of alternative hypothesis.
For \alg{Asymp-TW}, we only use the first graph in the sub-group for testing and use $r=10$ communities for approximation.
We run the above setup for 1000 independent trials (the randomness is induced by the splits of the classes during network construction) and report the powers of both tests in Tables~\ref{tab_seiz_pow2} and~\ref{tab_seiz_pow1}.

Table~\ref{tab_seiz_pow2} shows that for {\bf G$i$.1} vs. {\bf G$i$.2}, the null hypothesis is nearly always accepted by \alg{Asymp-Normal} (rejection rate less than 1.1\%).  In other cases, the rejection is 100\% except for {\bf G2.$x$} vs. {\bf G3.$x$}
which shows that these two classes have identical behaviour.
Table~\ref{tab_seiz_pow1} shows that  \alg{Asymp-TW} arrives at mostly similar conclusions, but in several cases of alternative hypothesis the power can be much smaller than 1. This is not surprising since the problem is harder for $m=1$.
We note that the authors of the dataset also do not claim that the various rest states can be distinguished, and state that the data is often used for binary setting of Group-1 (seizure) against other rest states.
To this end, both tests clearly show that Group-1 is significantly different from all other groups (100\% rejection).

A surprising observation from Table~\ref{tab_seiz_pow1} $(m=1)$ is that the rejection rate is 100\% within Group-1 ({\bf G1.1}	vs. {\bf G1.2}), whereas this is not the case for Table~\ref{tab_seiz_pow2} $(m=2)$.
This agrees with the conclusion of~\citet{Ghoshdastidar_2017_arxiv_00833} that the graph testing problem is fundamentally different for $m=1$ and $m>1$.
Our intuition is that the networks for seizure activity are significantly different from each other, and hence, rejected by \alg{Asymp-TW}.
When we group them $(m>1)$, the fundamental question is whether two groups are identically distributed or not, and hence, the variance within each group is also taken into account. Hence, \alg{Asymp-Normal} detects that {\bf G1.1} and {\bf G1.2} are identical when both graphs in each group are considered.

\begin{table}[t]
\centering
\caption{Negative logarithm of $\pv$ (averaged over 1000 runs) computed by \alg{Asymp-Normal} for EEG correlation networks.}
\label{tab_seiz_pval2}
\begin{tabular}{|c|cc|cc|cc|cc|cc|}
\hline
	&	{\bf G1.1}	&	{\bf G1.2}	&	{\bf G2.1}	&	{\bf G2.2}	&	{\bf G3.1}	&	{\bf G3.2}	&	{\bf G4.1}	&	{\bf G4.2}	&	{\bf G5.1}	&	{\bf G5.2}	\\
\hline
{\bf G1.1}	&	0	&	0.7	&	47.3	&	47.5	&	63.6	&	63.5	&	176.2	&	176.1	&	37.6	&	37.5	\\
{\bf G1.2}	&	0.7	&	0	&	47.4	&	47.7	&	63.7	&	63.6	&	176.5	&	176.5	&	37.9	&	37.8	\\
\hline
{\bf G2.1}	&	47.3	&	47.4	&	0	&	0.5	&	1.0	&	1.0	&	331.8	&	332.0	&	37.2	&	37.0	\\
{\bf G2.2}	&	47.5	&	47.7	&	0.5	&	0	&	1.0	&	1.0	&	331.6	&	331.9	&	37.1	&	37.1	\\
\hline
{\bf G3.1}	&	63.6	&	63.7	&	1.0	&	1.0	&	0	&	0.2	&	407.5	&	407.7	&	61.8	&	61.9	\\
{\bf G3.2}	&	63.5	&	63.6	&	1.0	&	1.0	&	0.2	&	0	&	407.3	&	407.6	&	61.6	&	62.0	\\
\hline
{\bf G4.1}	&	176.2	&	176.5	&	331.8	&	331.6	&	407.5	&	407.3	&	0	&	0.3	&	45.7	&	45.3	\\
{\bf G4.2}	&	176.1	&	176.5	&	332.0	&	331.9	&	407.7	&	407.6	&	0.3	&	0	&	45.8	&	45.4	\\
\hline
{\bf G5.1}	&	37.6	&	37.9	&	37.2	&	37.1	&	61.8	&	61.6	&	45.7	&	45.8	&	0	&	0.6	\\
{\bf G5.2}	&	37.5	&	37.8	&	37.0	&	37.1	&	61.9	&	62.0	&	45.3	&	45.4	&	0.6	&	0	\\
\hline
\end{tabular}
\end{table}

\begin{table}[b]
\centering
\caption{Negative logarithm of $\pv$ (averaged over 1000 runs) computed by \alg{Asymp-TW} for EEG correlation networks.}
\label{tab_seiz_pval1}
\begin{tabular}{|c|cc|cc|cc|cc|cc|}
\hline
	&	{\bf G1.1}	&	{\bf G1.2}	&	{\bf G2.1}	&	{\bf G2.2}	&	{\bf G3.1}	&	{\bf G3.2}	&	{\bf G4.1}	&	{\bf G4.2}	&	{\bf G5.1}	&	{\bf G5.2}	\\
\hline
{\bf G1.1}	&	0	&	7.727	&	7.727	&	7.727	&	7.727	&	7.727	&	7.727	&	7.727	&	7.727	&	7.727	\\
{\bf G1.2}	&	7.727	&	0	&	7.727	&	7.727	&	7.727	&	7.727	&	7.727	&	7.727	&	7.727	&	7.727	\\
\hline
{\bf G2.1}	&	7.727	&	7.727	&	0	&	0.017	&	0.003	&	0.008	&	7.727	&	7.727	&	1.791	&	1.924	\\
{\bf G2.2}	&	7.727	&	7.727	&	0.017	&	0	&	0	&	0.009	&	7.727	&	7.727	&	1.841	&	1.928	\\
\hline
{\bf G3.1}	&	7.727	&	7.727	&	0.003	&	0	&	0	&	0	&	7.727	&	7.727	&	1.718	&	1.780	\\
{\bf G3.2}	&	7.727	&	7.727	&	0.008	&	0.009	&	0	&	0	&	7.727	&	7.727	&	1.823	&	1.889	\\
\hline
{\bf G4.1}	&	7.727	&	7.727	&	7.727	&	7.727	&	7.727	&	7.727	&	0	&	0.195	&	5.149	&	4.950	\\
{\bf G4.2}	&	7.727	&	7.727	&	7.727	&	7.727	&	7.727	&	7.727	&	0.195	&	0	&	4.821	&	4.952	\\
\hline
{\bf G5.1}	&	7.727	&	7.727	&	1.791	&	1.841	&	1.718	&	1.823	&	5.149	&	4.821	&	0	&	0.366	\\
{\bf G5.2}	&	7.727	&	7.727	&	1.924	&	1.928	&	1.780	&	1.889	&	4.950	&	4.952	&	0.366	&	0	\\
\hline
\end{tabular}
\end{table}

Although Tables~\ref{tab_seiz_pow2} and~\ref{tab_seiz_pow1} show that the different groups are typically rejected, they do not clearly show the degree of dissimilarity between two groups. The dissimilarity can quantified in terms of the $\pv$ obtained from the tests.
While $\pv\leq 5\%$ leads to rejection, we find that in many cases, the $\pv$ is exponentially small.
In Tables~\ref{tab_seiz_pval2} and~\ref{tab_seiz_pval1}, we show the negative logarithm of $\pv$, that is $-\ln(\pv)$, obtained from \alg{Asymp-Normal} and \alg{Asymp-TW}, respectively. 
The reported value is the average over 1000 independent runs.
We note that the 5\% significance level corresponds to $-\ln(\pv)\approx 3$, and hence, values larger than 3 correspond to rejection.
Table~\ref{tab_seiz_pval2} shows that this quantity can be as high as 400, and in particular, it shows that Group-4 is most dissimilar from other groups.
The results of Table~\ref{tab_seiz_pval1} are less conclusive since the maximum reported dissimilarity is only 7.727.
This is caused by our use of a pre-computed table for the Tracy-Widom distribution that does not return values arbitrarily close to 1 (see Appendix~\ref{sec_algo} for a discussion provided along with the description of the test).
However, Table~\ref{tab_seiz_pval1} still shows that the networks in Group-5 are relatively less different from those in Groups-2, 3 and 4.

\subsection{Experiments with autonomous systems peering networks}
\label{sec_oregon}

Our second experiment with real networks is based on a collection of networks obtained from the Stanford large network collection~\citep{Leskovec_2014_data_snap}.
The networks are defined on the set of autonomous systems, which is the technical term for groups of routers that comprise the Internet.
The edges correspond to communication between two autonomous systems.
The first set of networks, called Oregon-1, are created from data collected by \emph{Oregon route-views} between March 31, 2001 and May 26, 2001 once per week. This set contains 9 networks, one date per week.
The second set of networks, called Oregon-2, are based on data collected on the same dates, but the peering information is inferred from a combination of \emph{Oregon route-views, Looking glass, and Routing registry}.

All the networks are defined on a set of $n=11806$ distinct vertices (autonomous systems), but none of the networks include all vertices, that is, every graph has few isolated vertices.
The networks are also quite sparse with the number of edges varying between 22000  to 33000.
We view the network collection from the following perspective.
For each date, we observe two networks (one from each set) that can be considered as a population of size 2 $(m=2)$.
Different dates correspond to different models for the networks, and we test for the similarity across different classes.
To this end, we perform \alg{Asymp-Normal} to detect differences, and report $-\ln(\pv)$ for every test in Table~\ref{tab_oregon}.
It is not surprising to find that the test rejects the null hypothesis at 5\% significance for every pair of dates, that is, $-\ln(\pv)>3$.
The interesting observation is that $-\ln(\pv)$ monotonically increases as the interval between two dates becomes larger, that is,
the networks vary significantly over time. 
This observation is also in conjunction with the findings of~\citet{Leskovec_2005_conf_KDD}, where a more qualitative analysis was made based on number of edges and average node degree.
We do not report corresponding results for \alg{Asymp-TW} since our current implementation can provide a maximum $-\ln(\pv)$ of at most 7.727, and hence, does not provide any additional information.

\renewcommand{\arraystretch}{1.15}

\begin{table}[b]
\centering
\caption{Negative logarithm of $\pv$ obtained by \alg{Asymp-TW} for every pair of dates in the Oregon network dataset.}
\label{tab_oregon}
\begin{tabular}{|c|@{}ccccccccc@{}|}
\hline
& Mar 31 & Apr 7 & Apr 14 & Apr 21 & Apr 28 & May 5 & May 12 & May 19 & May 26\\
\hline
Mar 31 &        0  & 13.7 &  25.0 & 36.4  & 59.6 &  77.4  & 96.8 &106.2 & 135.0 \\
Apr 7   &   13.7  &      0 &   6.5  & 15.2  & 31.0 &  45.7  & 61.1 & 69.7  & 93.4  \\
Apr 14 &   25.0  &  6.5  &       0 &   6.0  & 17.9 &  29.6  & 42.5 &  50.2 &  71.4 \\
Apr 21 &   36.4  & 15.2 &   6.0  &      0  &  8.5  & 17.2   & 27.6 &  34.9 &  54.7 \\
Apr 28 &   59.6  & 31.0 &  17.9 &   8.5  &     0   & 5.3    & 12.8 & 22.6  & 45.7  \\
May 5 &   77.4  & 45.7 &  29.6 &  17.2 &   5.3  &     0   & 4.8   & 13.0  & 31.2  \\
May 12 &   96.8  & 61.1 &  42.5 &  27.6 &  12.8 &  4.8   &     0  &  4.7   & 18.3  \\
May 19 &  106.2 & 69.7 &  50.2 &  34.9 &  22.6 &  13.0 & 4.7   &     0   & 5.6   \\
May 26 &  135.1 & 93.4 &  71.4 &  54.7 &  45.7 &  31.2 & 18.3 &  5.6   &     0 \\
\hline
\end{tabular}
\end{table}

\begin{figure}[t]
\centering
\begin{tabular}{cl}
\rotatebox{90}{\hskip8ex $-\ln(\pv)$} & 
\includegraphics[width=0.8\textwidth]{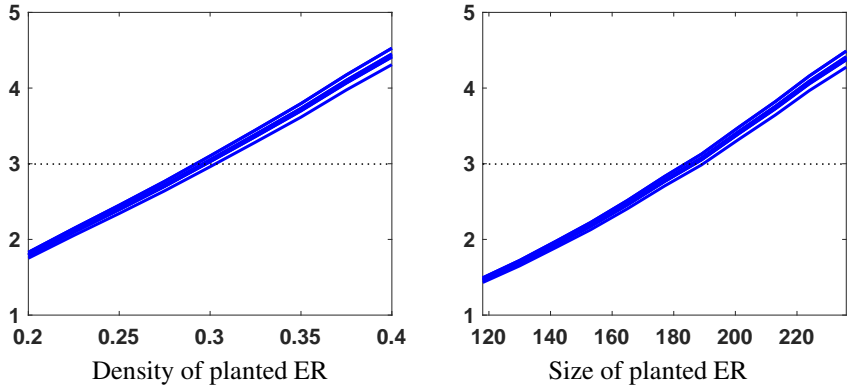}\\
& \hskip7ex Density of planted ER \hskip18ex Size of planted ER \\
\end{tabular}
\caption{Variation of $-\ln(\pv)$ for \alg{Asymp-Normal} when Erd\H{o}s-R{\'e}nyi subgraphs are planted into the network. Each line corresponds to one of the 9 pairs. The dotted line corresponds to 5\% significance level. {\bf(Left)} Subgraph size is 1\% of network size, and edge probability is varied. {\bf(Right)} Subgraph size is varied from 1-2\% of network size, and edge probability is decreased.}
\label{fig_oregon1}
\end{figure}

We next perform semi-synthetic experiments with Oregon network dataset.
We first consider the case of $m=2$, where we use \alg{Asymp-Normal}.
For every pair of networks, we randomly select $k=118$ vertices (1\% of vertex set), and replace the sub-graph by an Erd\H{o}s-R{\'e}nyi (ER) graph with edge probability $p$. On Figure~\ref{fig_oregon1} (left panel),  we show how $-\ln(\pv)$ varies as the edge density of the ER graph increases from $p=0.2$ to $0.4$, where each line corresponds to one date (one pair of networks) and the results are averaged over 100 runs.
We find that $-\ln(\pv)$ increases linearly with $p$, that is, $\pv$ decreases exponentially.
The trend is almost similar for every network pair.
We also study the effect of adding sparse ER graphs in Figure~\ref{fig_oregon1} (right panel).
Here we plant an ER graph on a random subset of $k$ vertices, where $k$ varies from 1\% to 2\% of total number of vertices.
However, the planted ER graphs are sparse with $p = \frac{20}{k}$, that is they have constant average degree.
We observe a slightly super-linear increase of  $-\ln(\pv)$  in this case.

\begin{figure}[b]
\centering
\begin{tabular}{cl}
\rotatebox{90}{\hskip8ex $-\ln(\pv)$} & 
\includegraphics[width=0.8\textwidth]{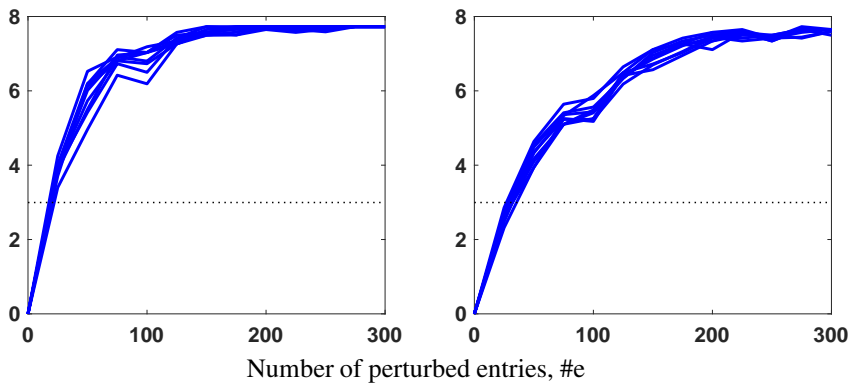}\\
& \hskip20ex Number of perturbed entries, \#e \\
\end{tabular}
\caption{Variation of $-\ln(\pv)$ for \alg{Asymp-TW} when a random set of \#e out of $\binom{n}{2}$ edges are inserted/deleted. The dotted line corresponds to 5\% significance level. {\bf(Left)} Each line corresponds to one of the 9 networks from Oregon-1 set. {\bf(Right)} Each line is for a network from Oregon-2 set.}
\label{fig_oregon2}
\end{figure}

Finally, we consider a semi-synthetic experiment with $m=1$, where we use \alg{Asymp-TW}.
For each of the 18 networks, we randomly select \#e pairs of vertices and toggle their connection, that is, if an edge is present then we remove it, or the reverse.
We vary \#e from 0 to 300 in steps of 25.
Figure~\ref{fig_oregon2} reports the values for $-\ln(\pv)$ (averaged over 100 runs) for each network.
We present the results in two panels corresponding to the two  datasets Oregon-1 and Oregon-2.
Surprisingly, we find that $-\ln(\pv)$ rapidly increases with \#e although the number of perturbed edges are much smaller than the total of $\binom{11806}{2}$ possible pairs.
We also find that the networks in each collection have a similar trend, and the Oregon-2 networks show a slightly smaller value than Oregon-1.
This is possibly because the Oregon-2 networks are  more dense than their Oregon-1 counterparts.

We conclude our discussion with some implementation details for \alg{Asymp-TW} in this setup related to the community detection step.
Since the networks are large and sparse, standard spectral clustering fails to return reasonable communities.
Hence, we use BigClam~\citep{Yang_2013_conf_WSDM}, which is suitable for finding a large number of communities in a large network.
The method returns multiple community assignments for some vertices and does not make any assignments for few.
We use BigClam to find an initial set of 50 overlapping communities from the union of all graphs, and then resolve cases of overlap or no-assignments by assigning vertices to communities with which they have maximum connection.
These pre-computed communities are used for the purpose of approximation in \alg{Asymp-TW} test.
The above results in Figure~\ref{fig_oregon2} show that the use of \alg{Asymp-TW} in conjunction BigClam provides reliable results, and hence, we believe that \alg{Asymp-TW} is applicable even in the sparse regime provided that it is used with a reasonable community detection algorithm.

\end{document}